\documentclass[preprint,12pt]{elsarticle}

\usepackage{amssymb}
\usepackage{caption}
\usepackage{subcaption}
\usepackage[flushleft]{threeparttable}
\usepackage{float}
\usepackage{amsmath}
\usepackage{graphicx}
\usepackage[ruled,vlined]{algorithm2e}
\usepackage{nomencl}

\usepackage[flushleft]{threeparttable}
\usepackage[numbers]{natbib}
\usepackage{xcolor}

\journal{Neurocomputing}

\begin{document}

\begin{frontmatter}



\title{LIFT-SLAM: a deep-learning feature-based monocular visual SLAM method}


\author[label1]{Hudson Martins Silva Bruno\corref{cor1}}
\ead{hudson.bruno@ic.unicamp.br}
\author[label1]{Esther Luna Colombini}

\address[label1]{Laboratory of Robotics and Cognitive Systems, Institute of Computing, State University of Campinas, Campinas, São Paulo, Brazil}
 
\begin{abstract}
The Simultaneous Localization and Mapping (SLAM) problem addresses the possibility of a robot to localize itself in an unknown environment and simultaneously build a consistent map of this environment. Recently, cameras have been successfully used to get the environment's features to perform SLAM, which is referred to as visual SLAM (VSLAM). However, classical VSLAM algorithms can be easily induced to fail when either the motion of the robot or the environment is too challenging. Although new approaches based on Deep Neural Networks (DNNs) have achieved promising results in VSLAM, they still are unable to outperform traditional methods. To leverage the robustness of deep learning to enhance traditional VSLAM systems, we propose to combine the potential of deep learning-based feature descriptors with the traditional geometry-based VSLAM, building a new VSLAM system called LIFT-SLAM. Experiments conducted on KITTI and Euroc datasets show that deep learning can be used to improve the performance of traditional VSLAM systems, as the proposed approach was able to achieve results comparable to the state-of-the-art while being robust to sensorial noise. We enhance the proposed VSLAM pipeline by avoiding parameter tuning for specific datasets with an adaptive approach while evaluating how transfer learning can affect the quality of the features extracted.
\end{abstract}




\end{frontmatter}


\section{Introduction}
\label{sec:introduction}

The ability to know its localization in an environment is an essential task for mobile robots, and it has been a subject of research in robotics for decades. To correctly localize itself, the robot must know its pose (position and orientation) in the environment. The process that estimates this information is called Odometry. When the robot simultaneously localizes itself and constructs a map of the unknown environment, the algorithm is called Simultaneous Localization and Mapping (SLAM).

In the last decades, the advances in hardware technologies, such as embedded GPUs, allowed significant advances on mobile robots pose estimation through camera-based methodologies of odometry and SLAM, which are called Visual Odometry (VO) and Visual SLAM (VSLAM). Much work has been done to develop accurate and robust VO and VSLAM systems. However, traditional approaches still depend on significant engineering effort on a classic pipeline: Initialization, feature detection, feature matching, outlier rejection, motion estimation, optimization, and relocalization. Furthermore, the traditional approaches tend to fail in challenging environments (inadequate illumination, featureless areas, etc.), when the camera is moving at high speed or if the camera suffers some distortions (rolling shutter effect, unfavorable exposure conditions, etc.). Moreover, if the camera is monocular, these systems have scale uncertainty.

Recently, many works have proposed using Deep Neural Networks (DNNs) to estimate camera motion with an end-to-end system. These systems can replace the entire traditional VO/VSLAM pipeline, which depends on significant engineering effort to develop and tune \cite{undeep-vo, attention-based, deep-vo}. However, these methods are not able to outperform traditional methods yet. Thus, some new works propose to replace only some modules of the VO/VSLAM traditional pipeline with DNNs, creating hybrid methods \cite{self-improving-vo, df-slam, pose-graph-optimization, gcnv2}. These approaches can leverage the robustness of deep learning to enhance traditional VSLAM systems. However, the literature still lacks an in-depth evaluation of these algorithms' robustness in challenging situations. Also, most of the proposed methods do not provide results in different scenarios to confirm the algorithms' robustness in all situations.

Therefore, in this paper, we propose employing the Learned Invariant Feature Transform (LIFT) \cite{lift} to extract features from images and use these features in a traditional VSLAM pipeline based on ORB-SLAM \cite{orb-slam} for monocular camera applications. Hence, we explore the potential of deep neural networks to improve the performance of conventional VSLAM systems. We also propose a set of experiments to evaluate the robustness of the algorithm in several scenarios. The main contributions of this work are summarized as follows:
\begin{itemize}
    \item This paper presents a novel hybrid VSLAM algorithm based on the LIFT network to perform feature extraction in a traditional back-end based on ORB-SLAM's system;
    
    \item We evaluate how transfer learning and fine-tuning can affect the quality of the resulting Hybrid VSLAM system;
    
    \item We extend the proposed system with an adaptive approach that can enhance its performance while avoiding fine-tuning of parameters that are usually dependable on the dataset;
    
    \item We conduct experiments on public KITTI \cite{kitti-dataset} and Euroc \cite{euroc-mav} datasets and present a set of experiments to confirm the robustness of algorithms based on learned features under camera distortions.
\end{itemize}
\section{Related Work}

\textbf{Feature-based approaches.} The first monocular feature-based VSLAM system proposed was MonoSLAM \cite{monoslam}. In this method, camera motion and 3D structure of an unknown environment are simultaneously estimated using an extended Kalman filter (EKF). There is no loop closure detection in this method, and it performs map initialization using a known object. This method's main problem is the computational cost, as it increases in proportion to the size of an environment. The Parallel Tracking and Mapping (PTAM) \cite{ptam} algorithm was proposed to solve the problems of MonoSLAM. To reduce computational costs, the authors propose to split tracking and mapping into two separate tasks, processed in parallel threads. That way, the tracking estimates camera motion in real-time, and the mapping estimates accurate 3D positions of feature points with a computational cost \cite{survey-visual-slam}. It is the first real-time method that was able to incorporate bundle adjustment (BA). They have also created an automatic initialization with a 5-point algorithm. The main ideas of PTAM were used in ORB-SLAM. 

Mur-Artal et al. proposed ORB-SLAM \cite{orb-slam}, a feature-based monocular VSLAM system with three threads: Tracking, Local Mapping, and Loop Closing. It relies on Oriented FAST and rotated BRIEF (ORB) features and uses a place recognition system based on Bag-of-Words (BoW). The mapping step adopts graph representations, which allow the system to perform local and global pose-graph optimization. Later, the authors of ORB-SLAM proposed an extension of ORB-SLAM applied to stereo and RGB-D cameras \cite{orb-slam2}. This is currently one of the state-of-the-art feature-based monocular VSLAM algorithms. However, because it is based on traditional features, ORB-SLAM can still fail in some situations, as shown in \cite{gcnv2}, requiring parameter tuning per dataset, as we will demonstrate later.


\textbf{End-to-end deep learning-based approaches.} One of the most notable end-to-end approaches is called DeepVO \cite{deep-vo}. In DeepVO, a Recurrent Neural Network (RNN) estimates the camera pose from features learned by a Convolutional Neural Network (CNN). The CNN architecture proposed is based on an architecture used to compute optical flow from a sequence of images called Flownet \cite{flownet}. Then two stacked Long-Short Term Memory (LSTM) layers are applied to estimate temporal changes from the features predicted by a CNN. Another end-to-end approach, based on unsupervised learning called UnDeepVO, is presented in \cite{undeep-vo}. The network relies on stereo image pairs to recover the scale during training while using consecutive monocular images for testing. Moreover, the loss function defined for training the networks uses spatial and temporal dense information. The system successfully estimates the pose of a monocular camera and the depth of its view. 

In \cite{attention-based}, an end-to-end system that uses a similar architecture to DeepVO is proposed. However, instead of employing LSTMs, they include an attention phase, which is called Neural Graph Optimization. It considers that poses that are temporally adjacent should have similar outputs and should be visually identical. Still, temporally different poses should also have related outputs, enabling a loop closure-like correction of drift. Although these approaches presented promising results, they are still not accurate enough to overcome the results of traditional methods.

\textbf{Hybrid approaches.} Hybrid approaches replace some modules of the traditional VSLAM pipeline. In \cite{pose-graph-optimization}, Li et. al. proposed a monocular system called Neural Bundler. It is an unsupervised DNN that estimates motion. Then, it constructs a conventional pose graph, enabling an efficient loop closing procedure based on the pose graph's optimization. A recent hybrid approach called SuperGlue \cite{superglue} proposed a graph neural network with an attention mechanism to perform the matching between two sets of local features. They use the DNN between feature extraction and pose estimation, which they call a learnable "middle-end," as it lies between the front-end and back-end of a traditional VSLAM system. 

Recently, some papers proposed using locally learned features to replace the traditional local features such as ORB and Scale-invariant feature transform (SIFT) of VSLAM systems. In DF-SLAM \cite{df-slam}, the TFeat network \cite{tfeat} is used to create descriptors for features extracted from stereo images with the FAST corner detector. The feature descriptors are then used in a traditional VSLAM pipeline, based on ORB-SLAM2 \cite{orb-slam2}. A self-supervised approach called SuperPointVO is proposed in \cite{self-improving-vo}, where they combine a DNN based in SuperPoint \cite{superpoint} feature extractor as a VSLAM front-end with a traditional back-end, using the stability of keypoints in the images to aid in learning. These approaches can leverage deep learning-based methods robustly and still be as accurate as a traditional feature-based approach. However, none of these papers evaluate their methods' robustness in challenging scenarios or in multiple datasets with different characteristics.
\section{Proposed Method}
\label{sec:method}

Our proposed method is a deep-learning feature-based monocular VSLAM system called LIFT-SLAM. It reconstructs sparse maps that are graph-based and keyframe-based, which allows us to perform bundle adjustment to optimize the estimated poses of the camera. We use the DNN called LIFT \cite{lift} to extract features that are used in a pipeline based on ORB-SLAM \cite{orb-slam}. 

\subsection{LIFT}
\label{sec:lift}
The Learned Invariant Feature Transform (LIFT) is a DNN proposed by Yi et al. \cite{lift} that implements local feature detection, orientation estimation, and description in a supervised end-to-end approach. The network architecture comprises three main modules based on CNNs: Detector, Orientation Estimator, and Descriptor.

\begin{figure}
\centering
\subfloat[LIFT pipeline.]{
\includegraphics[width=0.9\textwidth]{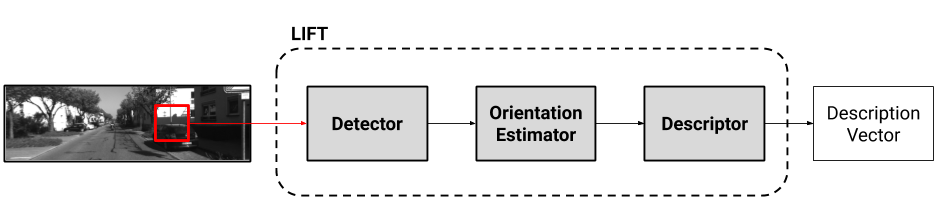}
\label{fig:lift}}

\subfloat[LIFT fine-tuned pipeline.]{
\includegraphics[width=0.9\textwidth]{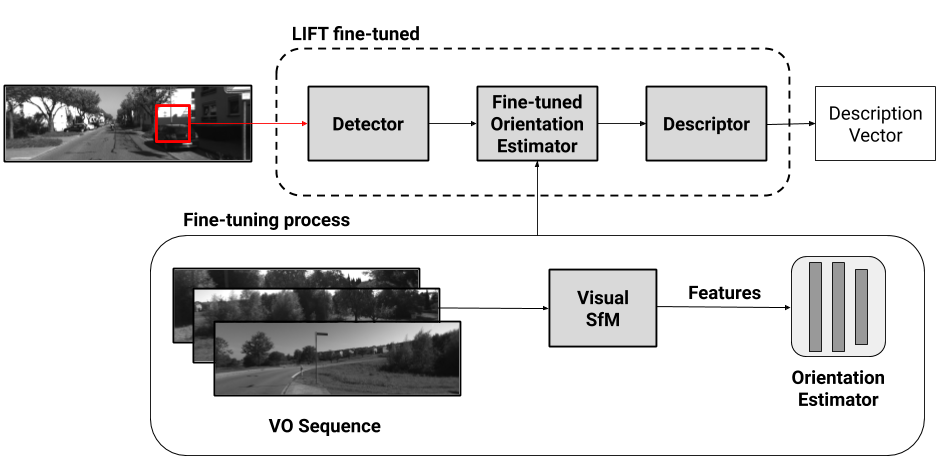}
\label{fig:lift-finetuned}}

\caption{LIFT and LIFT fine-tuned pipelines.}
\label{fig:lift-pipelines}
\end{figure}

The algorithm works with patches of images. After giving a patch as input, the detector network provides a score map of this patch. A soft argmax operation \cite{softargmax} is performed over this score map to return the potential feature point location. After this, it performs a crop operation centered on the feature location, used as input to the orientation estimator. The orientation estimator module predicts an orientation to the patch. Thus, a rotation is applied in the patch according to the estimated orientation. Lastly, the descriptor network computes a feature vector from the rotated patch, that is the output. This pipeline is presented in Figure \ref{fig:lift}.

Originally, LIFT was trained with photo-tourism image sets. They used a Structure from Motion (SfM) algorithm called VisualSFM \cite{visual-sfm} to reconstruct the scenes from the image sets with SIFT features. Photo-tourism data contains different geometrical aspects when compared to a typical VO dataset. Usually, in VO datasets, the images are sequential, captured with the same camera that progressively changes its position and orientation. On the other hand, the photo-tourism images capture views of the same scene from different perspectives. Therefore, to address this aspect, we perform a transfer learning in the LIFT network to generate a version of the LIFT that is fine-tuned with VO datasets' features. The only LIFT module that was improved after training the network was the orientation estimator, as shown in Figure \ref{fig:lift-finetuned}. 

\begin{figure}
\centerline{\includegraphics[scale=0.35]{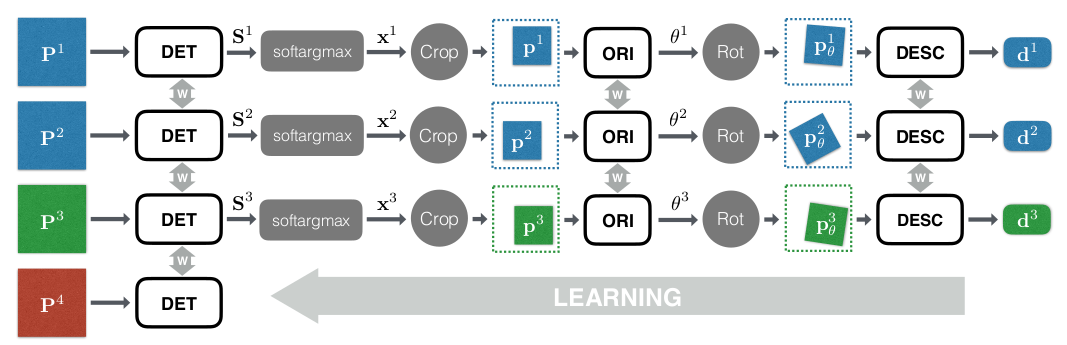}}
\caption[Siamese LIFT training architecture.]{The Siamese LIFT training architecture, composed of four branches. Where $\mathbf{P^i}$ is a patch of a image inputed on the $i$th branch, $S^i$ is a score map computed by the detector, $\mathbf{x^i}$ is a feature point location, and $\mathbf{p^i}$ is a smaller patch used as input to the orientation estimator. The orientation estimator computes a $\theta^i$ orientation that produces the rotated patch $\mathbf{p_{\theta}^i}$, this is processed by the descriptor network and produces a description vector $\mathbf{d^i}$. Extracted from \cite{lift}.}
\label{fig:lift-architecture}
\end{figure}

The LIFT training architecture is a four-branch Siamese, as shown in figure \ref{fig:lift-architecture}. In training, four patches of images are used as input, $\mathbf{P^1}$ and $\mathbf{P^2}$ correspond to different views of the same 3D point, $\mathbf{P_3}$ is a view from a different 3D point, and the last one $\mathbf{P^4}$ is a patch without any distinctive feature point. Each patch $\mathbf{P^i}$ corresponds to the $i$th branch of the network it will be used as input. The last branch trains only the detector network since it can only show negative examples to the detector.

The descriptor network is trained with a loss to minimize the differences between the corresponding patches and maximizing the difference between different patches. The descriptor is formalized as $h_{\rho}(\mathbf{p}_{\theta})$, where $\rho$ are the descriptor parameters. The descriptor is trained to minimize the loss defined in equation \ref{eq:desc-loss}.

\begin{equation}
    L_{desc}(\mathbf{p}_{\theta}^{k}) = \left\{\begin{matrix}
 \left \| h_{\rho}(\mathbf{p}_{\theta}^{k}) - h_{\rho}(\mathbf{p}_{\theta}^{l})  \right \|_2 & $for positive pairs$ \\ 
$max$(0, C-\left \| h_{\rho}(\mathbf{p}_{\theta}^{k})  -  h_{\rho}(\mathbf{p}_{\theta}^{l}) \right \|_2) & $for negative pairs,$ 
\end{matrix}\right.
    \label{eq:desc-loss}
\end{equation}

where C = 4, and positive pairs are patches that correspond to the same 3D point, and negative patches are the ones that do not correspond.

Moreover, the orientation estimator network is trained to provide the orientations that minimize the distances between description vectors for different views of the same 3D points. In training the orientation estimator, the description vectors are provided by the already trained descriptor, and the keypoints are taken from VisualSFM. The orientation estimator loss is defined in equation \ref{eq:ori-loss}.

\begin{equation}
    L_{ori}(\mathbf{P}^{1}, \mathbf{x}^{1}, \mathbf{P}^{2}, \mathbf{x}^2) =  \left \| h_{\rho}(G(\mathbf{P}^{1}, \mathbf{x}^{1}) - h_{\rho}(G(\mathbf{P}^{2}, \mathbf{x}^{2})) \right \|_2,
    \label{eq:ori-loss}
\end{equation}
where $G(\mathbf{P},\mathbf{x})$ is rotation applied to the patch $P$ centered in location $x$.

Finally, the detector learns to minimize the distance between the description vectors for corresponding patches (with the already learned descriptor and orientation estimator) and maximize the classification score for patches that do not correspond to the same physical point. Therefore, the loss in detector ($L_{det}$) is the sum of two losses $L_{class}$ and $L_{pair}$, as shown in equation \ref{eq:loss-det}. The detector output (score map) is defined as $\emph{f}_{\mu}(\mathbf{P})$, where $\mu$ are the network parameters.

\begin{equation}
    L_{det}(\mathbf{P}^{1}, \mathbf{P}^{2}, \mathbf{P}^{3}, \mathbf{P}^{4}) = \gamma L_{class}(\mathbf{P}^{1}, \mathbf{P}^{2}, \mathbf{P}^{3}, \mathbf{P}^{4}) + L_{pair}(\mathbf{P}^{1}, \mathbf{P}^{2}),
    \label{eq:loss-det}
\end{equation}
where $\gamma$ is a hyper-parameter that defines a balancing between the two terms, with $L_{class}$ increasing when detecting a keypoint in patch $\mathbf{P}^{4}$, as in equation \ref{eq:loss-class}. $L_{pair}$ is defined in equation \ref{eq:loss-pair}, it defines the distance between two corresponding description vectors.

\begin{equation}
    L_{class}(\mathbf{P}^{1}, \mathbf{P}^{2}, \mathbf{P}^{3}, \mathbf{P}^{4}) = \sum_{i=1}^{4} \alpha_{i} \textrm{max}(0, (1-\textrm{softmax}(\emph{f}_{\mu}(\mathbf{P}^i))y_{i}))^2,
\label{eq:loss-class}
\end{equation}
where $y_i = -1$ and $\alpha_i = 3/6$ if $i=4$ (for a non-keypoint patch), and $y_i = +1$ and $\alpha_i = 1/6$ otherwise. The \textrm{softmax} is a non-linear function.

\begin{equation}
\begin{split}
     L_{pair}(\mathbf{P}^{1}, \mathbf{P}^{2}) = \parallel &h_{\rho}(G(\mathbf{P}^{1}, \textrm{softargmax}(\emph{f}_{\mu}(\mathbf{P}^1)))) - \\& h_{\rho}(G(\mathbf{P}^{2}, \textrm{softargmax}(\emph{f}_{\mu}(\mathbf{P}^2))))\parallel_2,
\end{split}
\label{eq:loss-pair}
\end{equation}
where the \textrm{softargmax} is a function that computes the center of mass of the score map, returning the feature location $\mathbf{x}$.

Before constructing the pipeline of LIFT-SLAM, we tested the robustness of LIFT descriptors under different scenarios. To this end, we performed a qualitative analysis of the LIFT feature matching in sequential images from the KITTI dataset and compared them with ORB feature matching in the same conditions. First, we extract the features from a pair of images. Then, we find the descriptors' pair with a smaller distance between them (similarity) to create a match. LIFT descriptors are vectors of float numbers. Therefore, we compute the distance between the two descriptors with the Euclidean distance. On the other hand, ORB descriptors are binary vectors. Thus, we calculate the similarity between two ORB descriptors with Hamming distance.

We have created three different scenarios:
\begin{itemize}
    \item Frame skipping: To evaluate the feature matching performance, we emulate different camera frequencies by skipping frames in sequences of the KITTI dataset. Therefore, given an image in time $t$, we look for feature matches with an image in time $t+5$.
    
    \item Gamma power transformation with $\gamma > 1$: We can emulate under and overexposed images with gamma power transformation, as shown in \cite{emulate-exposure}. This transformation creates a new image $I'$ from image $I$ by applying: $I'=I^{\gamma}$. For $\gamma > 1$, we emulate an underexposed image.
    
    \item Gamma power transformation with $\gamma < 1$: We apply the same operation as before, but using $\gamma < 1$ to emulate an overexposed image.
\end{itemize}

\begin{figure}
\centering
\subfloat[LIFT feature matching in KITTI dataset after skipping 5 frames.]{
\includegraphics[width=0.9\textwidth]{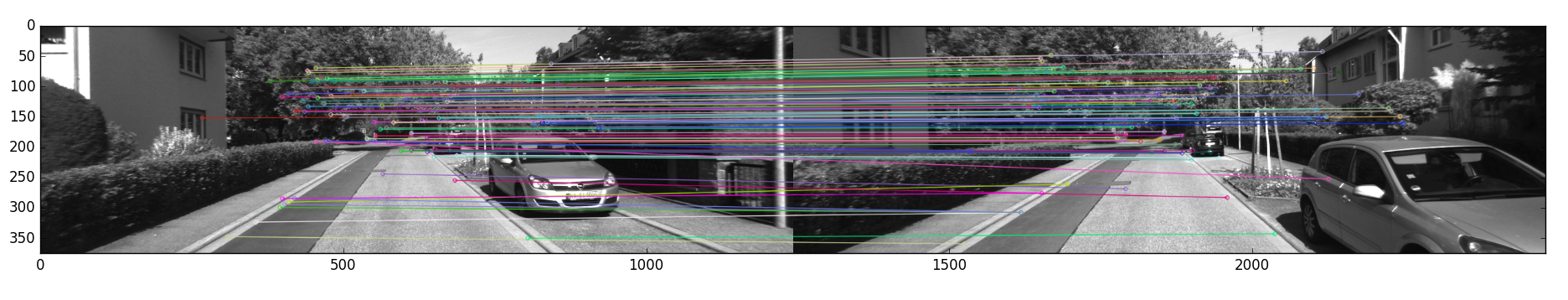}
\label{fig:lift-skip5}}

\subfloat[ORB feature matching in KITTI dataset after skipping 5 frames.]{
\includegraphics[width=0.9\textwidth]{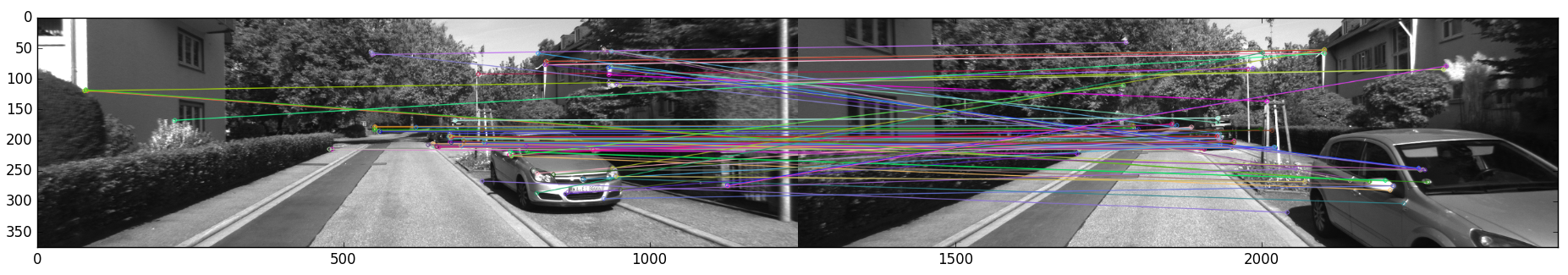}
\label{fig:orb-skip5}}

\subfloat[LIFT feature matching in KITTI dataset after gamma power transformation with $\gamma = 2$.]{
\includegraphics[width=0.9\textwidth]{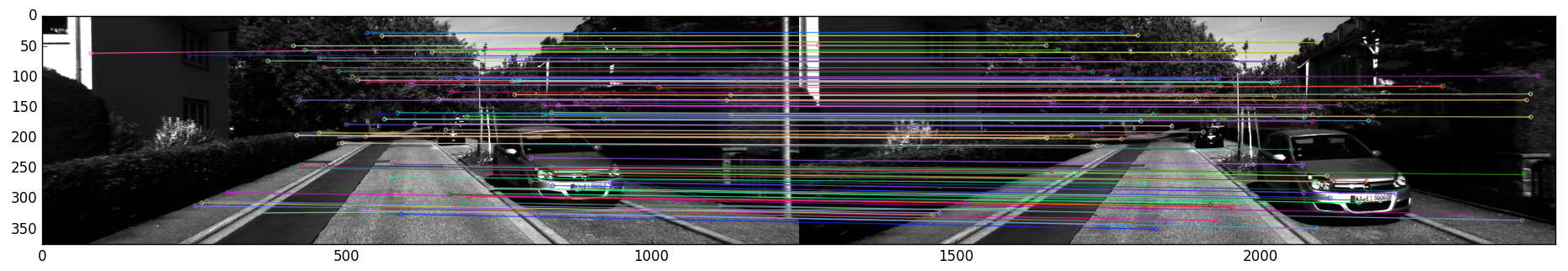}
\label{fig:lift-gamma2}}

\subfloat[][ORB feature matching in KITTI dataset after gamma power transformation with $\gamma = 2$.]{
\includegraphics[width=0.9\textwidth]{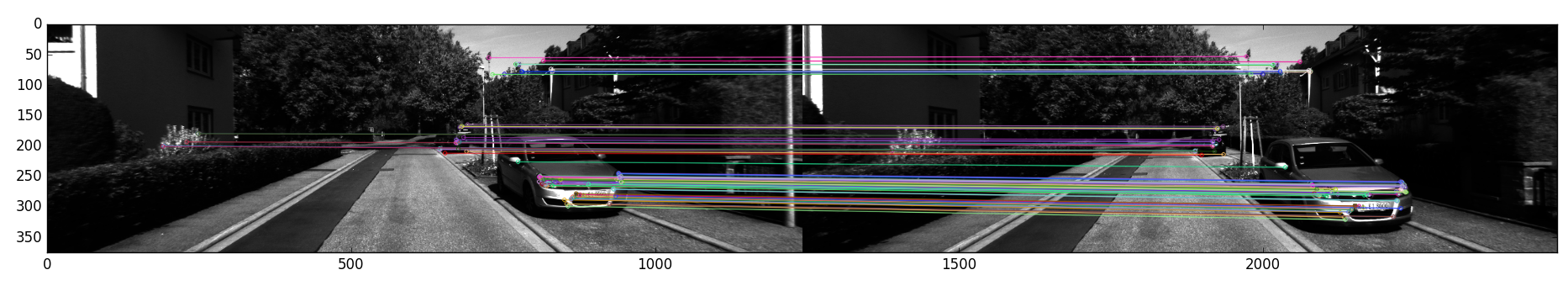}
\label{fig:orb-gamma2}}

\subfloat[LIFT feature matching in KITTI dataset after gamma power transformation with $\gamma = \frac{1}{2}$.]{
\includegraphics[width=0.9\textwidth]{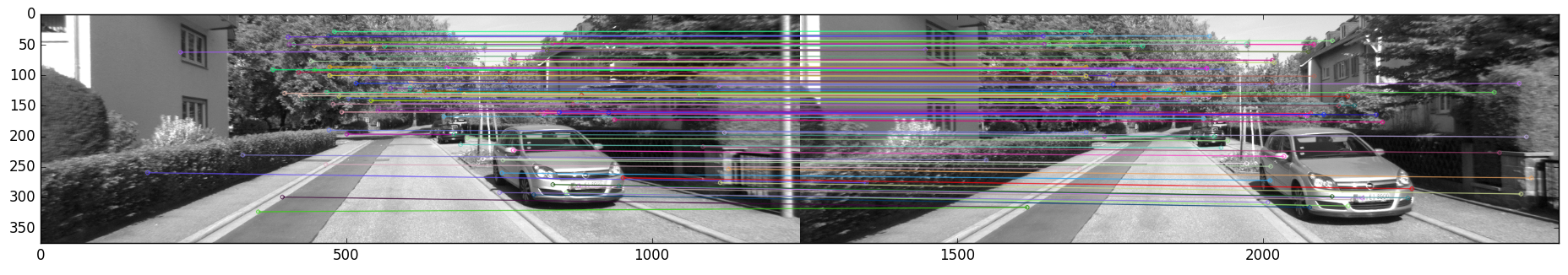}
\label{fig:lift-gamma0.5}}

\subfloat[ORB feature matching in KITTI dataset after gamma power transformation of $\gamma = \frac{1}{2}$.]{
\includegraphics[width=0.9\textwidth]{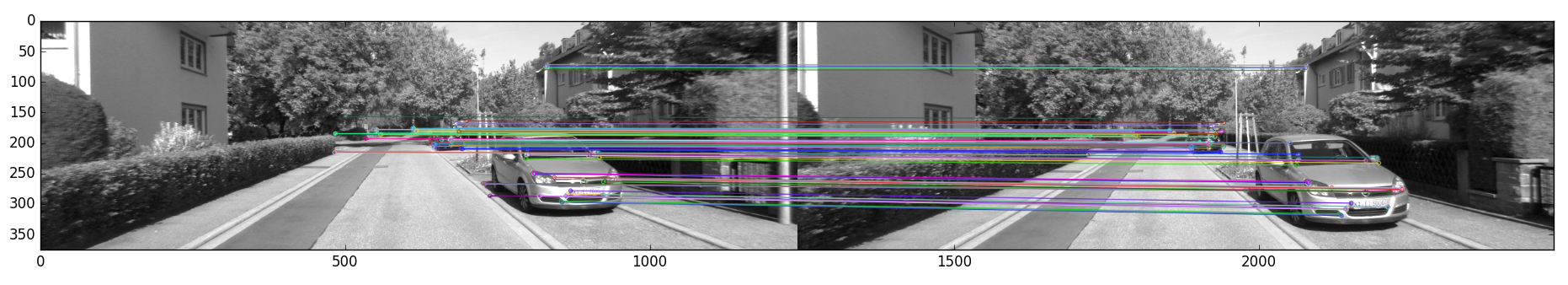}
\label{fig:orb-gamma0.5}}

\caption{Comparison between features LIFT and ORB under different conditions. Lines are connecting corresponding keypoints computed by feature matching. We show only the best 100 matches in each Figure.}
\label{fig:lift-orb-features}
\end{figure}

From the qualitative results shown in Figure \ref{fig:lift-orb-features}, we concluded that LIFT is robust to all of the proposed scenarios. In frame skipping (Figure \ref{fig:lift-skip5}), we can notice that LIFT is still able to find correct correspondences between images, whereas ORB creates several wrong correspondences (Figure \ref{fig:orb-skip5}). In gamma power transformations, the feature matching with both descriptors can create correct correspondences, however, the ORB keypoints are grouped in only a few regions of the images (Figures \ref{fig:orb-gamma2} and \ref{fig:orb-gamma0.5}). On the other side, the LIFT keypoints matched are spread within the whole image (Figures \ref{fig:lift-gamma2} and \ref{fig:lift-gamma0.5}), this aspect can improve accuracy of VO systems, as discussed in \cite{tutorial-vo-2}.

\subsection{LIFT-SLAM Pipeline}

\begin{figure}
\centerline{\includegraphics[scale=0.25]{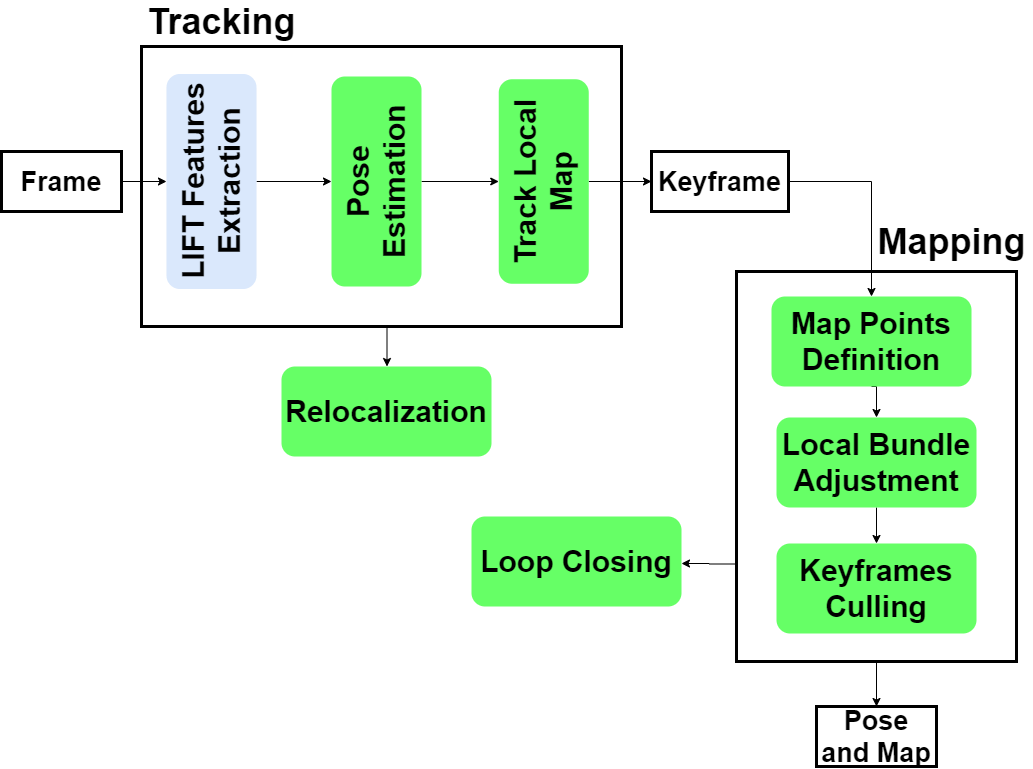}}
\caption[LIFT-SLAM pipeline.]{An overview of LIFT-SLAM pipeline, where tracking and mapping are sequential tasks, relocalization is called when the tracking of the camera pose is lost, and loop closing is a task that runs in parallel over the keyframes processed by mapping.}
\label{fig:lift-slam-pipeline}
\end{figure}

As aforementioned, our pipeline is very similar to the pipeline of ORB-SLAM \cite{orb-slam}. However, as we are not aiming, at this point, in an online version of the method, the mapping step runs sequentially after tracking and not in parallel, as in ORB-SLAM. Thus the only task we run in parallel is loop closure detection. Figure \ref{fig:lift-slam-pipeline} shows an overview of our pipeline that is described next.

\textbf{Tracking.} In tracking, for each frame, we extract LIFT keypoints and descriptors. We use these features in all feature matching operations needed in initialization, tracking, mapping, and place recognition. Then, as in ORB-SLAM, the camera pose is predicted with a constant velocity model. Later, we optimize the camera pose by searching for more map point correspondences in the current frame by projecting the local map 3D points into the image. Lastly, the tracking step decides if the current frame should be a keyframe.

\textbf{Mapping.} For each new keyframe, the mapping step is performed. First, it inserts the keyframe into the covisibility graph as a new node, and its edges are computed based on the shared map points with other keyframes. Furthermore, new map points are created by triangulating LIFT features from keyframes connected in the covisibility graph. A local bundle adjustment is responsible for optimizing the covisibility graph. It is applied to all keyframes connected to the current keyframe in the covisibility graph (including the current keyframe) and all map points seen by those keyframes. Finally, in keyframes culling, we discard keyframes that are redundant to improve the covisibility graph's size. This is useful since BA is a costly operation that grows in complexity as the number of keyframes increases. 

\textbf{Relocalization and loop closure.} To perform place recognition, we have created a visual vocabulary in an offline step with the DBoW2 library\footnote{https://github.com/dorian3d/DBoW2} \cite{dbow2}. The dictionary was created with LIFT descriptors of approximately 12,000 images collected from outdoors and indoor sequences from the TUM-mono VO dataset \cite{tum-mono-vo}. In this way, we can generate a vocabulary that provides good results in both environments. The built vocabulary has six levels and 10 clusters per level. Thus we get $10^6$ visual words, as suggested in \cite{vocabulary-based-slam}. If the tracking is lost, we query the Bag of Words (BoW) of the current frame into the database to find keyframe candidates for global relocalization.


The loop closing task runs in a separate thread. It gets the last keyframe processed by the local mapping and tries to detect if it closes a loop. After converting the keyframes to BoW, a similarity score between the current keyframe and its neighbors' covisibility graph is computed. The similarity between two BoW is given by the L2-score, as defined in \cite{vocabulary-tree}. The loop candidates are accepted if there are at least three candidates detected in the same covisibility graph. After finding the loop candidates, it computes a rigid-body transformation from the candidate keyframe to the loop keyframe. This transformation, the similarity transformation, informs about the drift accumulated in the trajectory, and it also works as a geometrical validation of the loop. If a similarity transformation is successfully found, we proceed to correct the loop. 


\subsection{Versions of LIFT-SLAM}
To explore the potential of our approach and to find changes that might lead to an improvement in general results, we developed some different versions of LIFT-SLAM. The next sections describe the decision process to create these versions and how we developed them.

\subsubsection{Fine-tuned LIFT-SLAM}
\label{sec:finetuned-lift}

In this version of LIFT-SLAM, we use these fine-tuned models to perform feature detection and description, as shown in Figure \ref{fig:lift-finetuned}. To refine the LIFT network, we had to collect the ground-truth data. As proposed in LIFT's paper \cite{lift}, we generate the ground-truth with SIFT keypoints collected with VisualSFM. We created two sets of ground-truth data. The first one comprises images from sequences 00, 06, 09, and 10 (8434 images) of the KITTI dataset, whereas the second contains images from the sequences MH\_04, V1\_03, and V2\_03 (6104 images) of the Euroc dataset. After collecting the datasets, we train the network in two versions, one for each dataset. We used the TensorFlow version of LIFT provided by the authors in their \textit{github}\footnote{All LIFT code used in this project comes from github.com/cvlab-epfl/tf-lift}. 

\subsection{Adaptive LIFT-SLAM}
A wrong data association in feature matching might affect the quality of the motion estimation. Therefore, to select the best features, a threshold is applied right after feature matching. In this way, the matches with greater distance than this threshold are discarded. On the other hand, if the threshold value is too small, we might reject good matches and loose track of the camera pose in challenging environments. We define two thresholds to mitigate this problem: the higher threshold ($TH_{HIGH}$) and the lower threshold ($TH_{LOW}$). We use $TH_{LOW}$ when we need to be more restrictive about the quality of the matches, as in relocalization or map point triangulation.

However, while performing our experiments, we found out that for different datasets, the best values for these thresholds could change, as shown in table \ref{tab:threshold-problem}. Therefore, we had to change the limits every time we needed to change the dataset. This is not desirable since, in real-world applications, it is not possible to deduce these thresholds' values. Hence, we have developed an adaptive method that decides the threshold values online, based on the number of outliers of the current frame and the number of map points in the last frame.

\begin{table}[!htb]
\centering
\resizebox{0.5\textwidth}{!}{\begin{tabular}{|cc|cc|}
\hline
\multicolumn{2}{|c|}{\textbf{Threshold}} & \multicolumn{2}{c|}{\textbf{ATE (m)}} \\\hline
\textbf{$TH_{LOW}$} & \textbf{$TH_{HIGH}$} & MH\_01 & KITTI 05  \\ \hline
1.0 & 2.0 & 0.052 & X \\ \hline
1.0 & 1.5 & 0.049 & X \\ \hline
2.0 & 3.0 & 0.629 &  12.61\\ \hline
\end{tabular}}
\caption{Absolute Trajectory Error (ATE) \cite{tum-vi} for different matching thresholds in KITTI and Euroc datasets. In Euroc MH\_01 sequence, the error is completely different for different thresholds, where the best thresholds are $TH_{LOW} = 1.0$ and $TH_{HIGH}=1.5$. Furthermore, in KITTI 05 sequence, the algorithm could track the camera pose only with $TH_{LOW} = 2.0$ and $TH_{HIGH}=3.0$.}
\label{tab:threshold-problem}
\end{table}

\begin{figure}
\centerline{\includegraphics[width=\textwidth]{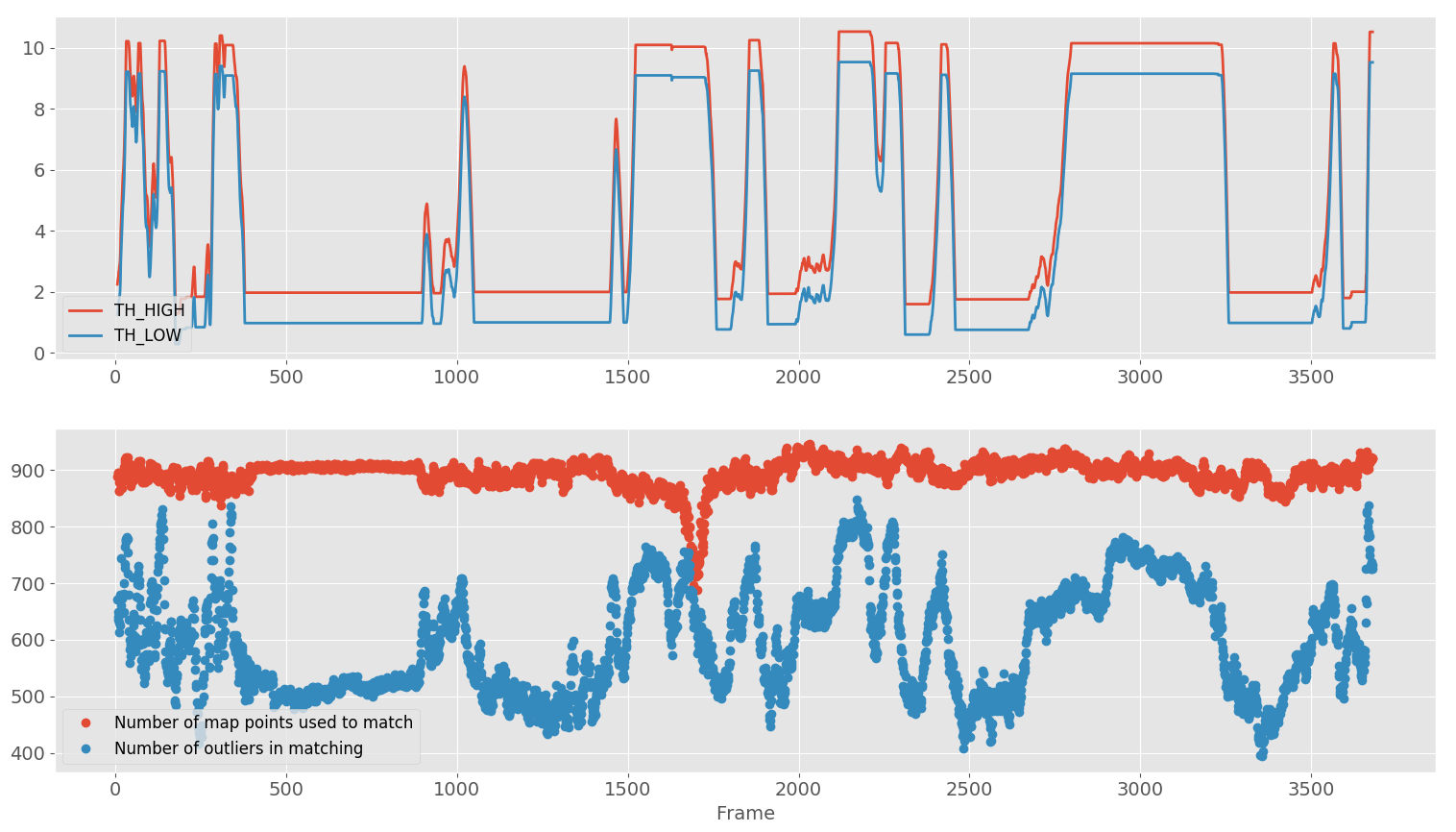}}
\caption[Thresholds in Adaptive LIFT-SLAM.]{Thresholds variation based on the number of map points used to match and the number of outliers after performing the matching. The threshold values decrease as the difference between the number of outliers and map points increases and vice versa.}
\label{fig:adaptive-thresholds}
\end{figure}

After estimating the pose with the constant velocity model, we search map point correspondences by projecting the map points from the last frame into the current frame. If the number of outliers gets approaches the number of map points, the number of matches gets too small and, consequently, the tracking is lost. We use this fact to create our adaptive method. It changes the thresholds values based on the difference between the number of map points and the number of outliers. Therefore, if this difference decrease, we increase the values of the thresholds. Figure \ref{fig:adaptive-thresholds} depicts the variation of the threshold based on the number of map points used to match and the number of outliers after performing the matching with the adaptive method used in this version of LIFT-SLAM. 

\section{Experiments}
\label{sec:experiments}

\subsection{Datasets}

Our experiments were performed in KITTI dataset \cite{kitti-dataset} and Euroc MAV Dataset \cite{euroc-mav}. KITTI dataset (Karlsruhe Institute of Technology and Toyota Technological Institute at Chicago) \cite{kitti-dataset} is one of the most used benchmarks for evaluation in VO/VSLAM algorithms. They have developed benchmarks for stereo, optical flow, VO/VSLAM, and 3D object detection. The VO/VSLAM dataset consists of 22 stereo images sequences with a total length of 39.2 km recorded from a moving car. As our goal is to work with monocular images, we get only the left images in all sequences to run our algorithms. Moreover, we use only sequences from $00$ to $10$, as these are the only sequences with ground-truth information available.

The Euroc MAV dataset (Swiss Federal Institute of Technology and Autonomous Systems Lab) \cite{euroc-mav} is a dataset created to assess the visual-inertial SLAM and 3D reconstruction capabilities of contestants from the European Robotics Challenge (Euroc) on Micro Aerial Vehicles (MAVs). Eleven sequences are provided in total, ranging from slow flights under good visual conditions to dynamic flights with motion blur and poor illumination. There are two types of sequences. The first type is from images taken in a realistic industrial scenario, recorded in a machine hall (sequences from MH\_01 to MH\_05). The second type is from images taken inside a Vicon motion capture system, with obstacles placed over the scene (sequences from V1\_01 to V1\_03).

These datasets were chosen to test the robustness of the proposed algorithms for different camera motion (e.g., acceleration, velocities, DoF, etc.) and environments (e.g., outdoors/indoors, size, illumination, etc.). Moreover, in LIFT-SLAM fine-tuned approaches, using different datasets has allowed us to validate the network's improvement for VO problems in general, instead of biasing the network for a single dataset.

\subsection{Trajectory Evaluation}
\label{sec:traj-evaluation}
We generate a quantitative and qualitative comparison between the estimated trajectories and the ground-truth data for each sequence of the datasets. The quantitative evaluation in KITTI sequences are based on Relative Pose Error (RPE) of translation and rotation, as described in \cite{kitti-benchmark}, and Absolute Trajectory Error (ATE), detailed in \cite{tum-vi}. Due to the stochastic nature of the algorithms, all of the quantitative metrics are an average of 5 executions. The estimates on Euroc sequences were evaluated only by ATE. 

ORB-SLAM's results were computed by our executions since, in ORB-SLAM's paper, an evaluation with RPE is not presented and does not provide results in the Euroc dataset. Furthermore, we present qualitative comparisons showing a 2-D plot of the trajectories. Moreover, for LIFT-SLAM versions that are not adaptative we set the matching thresholds values to $TH_{LOW} = 1$ and $TH_{HIGH} = 2$ for Euroc sequences and $TH_{LOW} = 2$ and $TH_{HIGH} = 3$ for KITTI sequences.

The quantitative comparison of all algorithms in the KITTI dataset presented in table \ref{tab:all-results-kitti} shows that, in general, LIFT-SLAM systems presented a better performance than ORB-SLAM, especially in smaller sequences, such as $03$ and $04$. Furthermore, we can notice that the proposed versions of LIFT-SLAM achieved a better performance than LIFT-SLAM in most of the sequences. The algorithm performance improved even when we used the Euroc dataset to fine-tune the LIFT network. Therefore, we confirmed that the network learned important features from VO datasets. 

\begin{table}[!h]
\centering
\resizebox{\textwidth}{!}{\begin{tabular}{|c|c|ccccccccccc|}
\hline \textbf{Algorithm} & \textbf{Metric} & \textbf{00} & \textbf{01} & \textbf{02} & \textbf{03} & \textbf{04} & \textbf{05} & \textbf{06} & \textbf{07} & \textbf{08} & \textbf{09} & \textbf{10}\\ \hline

                         & ATE (m) & 11.54& X & X & 15.13& 4.29 & \textbf{7.74} & 20.26 & 13.47 &\textbf{39.51} & \textbf{49.67} & 19.94\\
ORB-SLAM      & $RPE_{trans}$ (\%) & 4.46 & X & X & 9.75 & 3.71 & \textbf{3.35} & 8.11 & 7.43 & \textbf{12.16} & 26.51 & 8.65 \\
              &$RPE_{rot}$ (deg/m) & 3.28 & X & X & 2.78 & 2.15 & 3.57 & 2.88 & 3.58 &3.05 & 11.13 & 3.62 \\ \hline

          & ATE (m)            &18.77 & X & X & 1.10 & 0.40 & 8.09 & 18.47 & 4.03 & 80.97 & 59.88 & 31.84 \\
LIFT-SLAM & $RPE_{trans}$ (\%) & 6.71          & X & X & 0.87          & \textbf{2.10} & 4.46 & 7.76 & 2.51 & 27.63 & 20.65 & 10.08 \\
          & $RPE_{rot}$ (deg/m)& \textbf{2.20} & X & X & \textbf{0.34} & 0.65 & 2.58 & 2.49 & 3.60 & 2.10 & 2.12 & 2.25 \\  \hline

                               & ATE (m)            & - & X & \textbf{29.83} & 1.91 & \textbf{0.36} & 12.47 & - & \textbf{2.54} & 188.51 & - & -  \\
LIFT-SLAM fine-tuned with KITTI & $RPE_{trans}$ (\%) & - & X & 8.80 & 1.32 & 2.16 & 5.02          & - & \textbf{1.80} & 48.90 & - & - \\
                     & $RPE_{rot}$ (deg/m)& - & X & \textbf{2.11} & \textbf{0.34} & 0.52 & 2.43  & - & \textbf{2.67} & 2.11 & - & -  \\  \hline
          
                               & ATE (m)    & 9.84 & X & 34.23 & 0.97 & 0.42 & 11.50 & \textbf{16.58} & 3.98 &  82.61 & 54.91 & 30.34\\
LIFT-SLAM fine-tuned with Euroc & $RPE_{trans}$ (\%) & 3.49 & X & 9.84 & 0.86 & 2.22 & 5.35          & \textbf{7.05}         & 2.60 & 28.99 & \textbf{19.16} & 9.81\\
                               & $RPE_{rot}$ (deg/m)& 2.63 & X & 2.10 & 0.46 & 0.50 & \textbf{1.91} & \textbf{2.36} & 3.64 & \textbf{1.95} & \textbf{2.08} & 2.20\\  \hline

          & ATE (m)                     & 13.70 & X & 40.33 & \textbf{0.84} & 0.47 & 10.85 & 17.83 & 4.09 & 81.69 & 57.74 & \textbf{10.51} \\
Adaptive LIFT-SLAM & $RPE_{trans}$ (\%) & \textbf{2.64} & X & 11.54         & \textbf{0.78} & 2.22 & 5.49 & 7.50 & 2.67 & 28.49 & 19.28 & 4.96 \\
          & $RPE_{rot}$ (deg/m)         & 4.95 & X & 2.22 & 0.38 & 0.60 & 2.97 & 2.42 & 3.42 & 2.05 & 2.17 & \textbf{1.57} \\  \hline

                                & ATE (m)                    & - & X & 48.09 & 1.91 & 0.42 & 10.35 & - & 4.10 & 185.15 & - & -  \\
Adaptive LIFT-SLAM fine-tuned with KITTI & $RPE_{trans}$ (\%) & - & X & 9.57 & 1.29        & 2.11 & 4.64 & - & 2.64 & 47.20 & - & - \\
                               & $RPE_{rot}$ (deg/m)         & - & X & 2.43 &\textbf{0.34}& 0.57 & 2.93 & - & 3.51 & 2.00 & - & -  \\  \hline

                                        & ATE (m) & \textbf{8.06} & X & 40.04 & 2.23 & 0.51 & 13.55 & 30.38 & 3.63 & 184.43 & 59.62 & 29.87 \\
Adaptive LIFT-SLAM fine-tuned with Euroc & $RPE_{trans}$ (\%) & 3.18 & X & \textbf{8.73} & 1.46 & 2.22 & 6.09 & 12.24 & 2.42 & 47.10 & 19.91 & 9.72 \\
                                        & $RPE_{rot}$ (deg/m) & 2.99 & X & 2.49 & \textbf{0.34} & \textbf{0.48} & 3.11 & 2.91 & 4.02 & 2.02 & 2.14 & 2.24  \\  \hline
\end{tabular}}

\caption{Quantitative comparison of ORB-SLAM and all versions of LIFT-SLAM in the KITTI dataset. We fill with "X" the sequences unavailable due to tracking failure, and with "-" sequences, we do not execute the algorithm to avoid biased results. The smaller average in each metric is highlighted.}
\label{tab:all-results-kitti}
\end{table}

\begin{figure}
\centering
\subfloat[][KITTI 00]{
\includegraphics[width=0.44\textwidth, height=4cm]{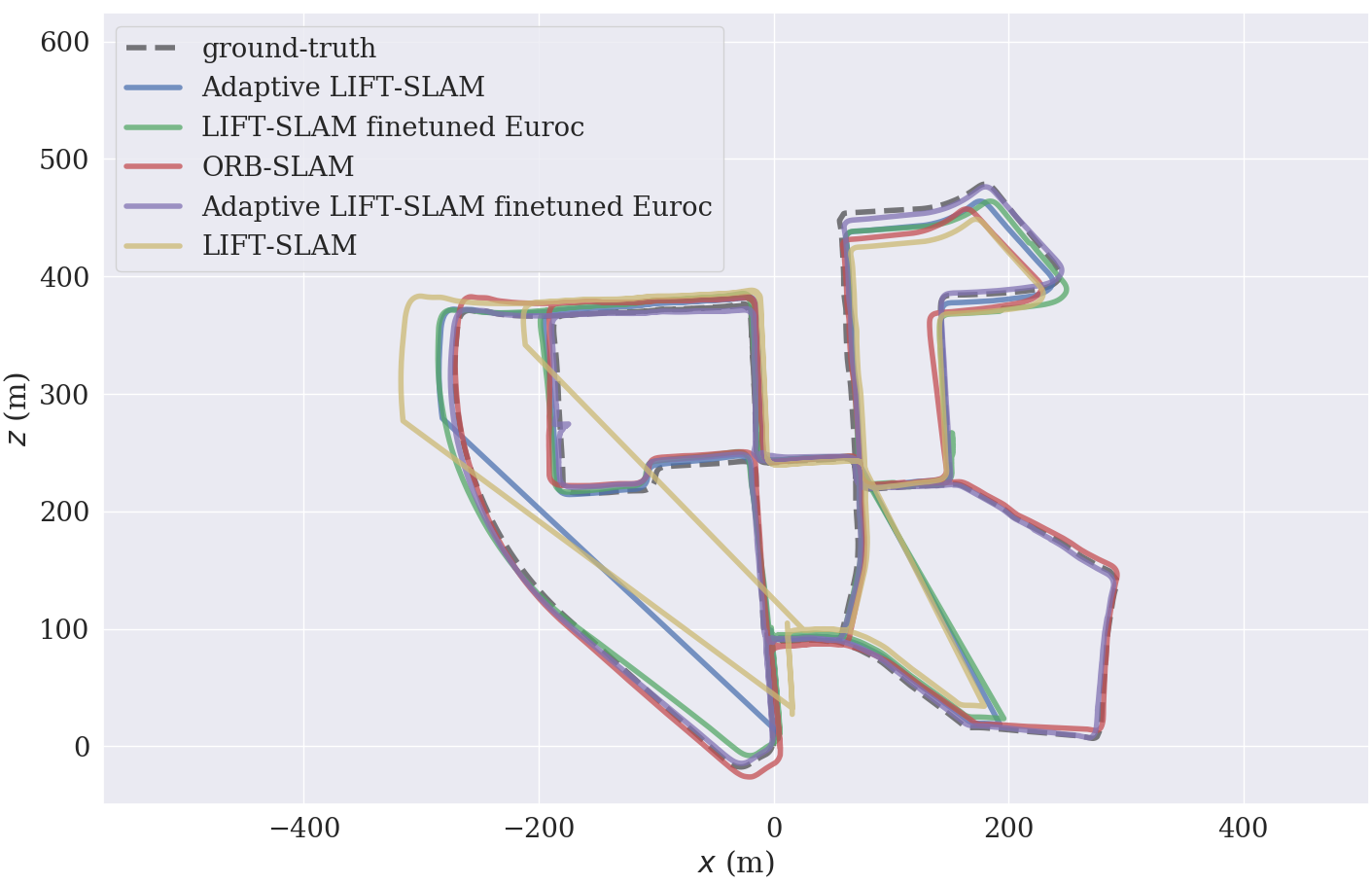}
\label{fig:kitti-00}}
\qquad
\subfloat[][KITTI 02]{
\includegraphics[width=0.44\textwidth, height=4cm]{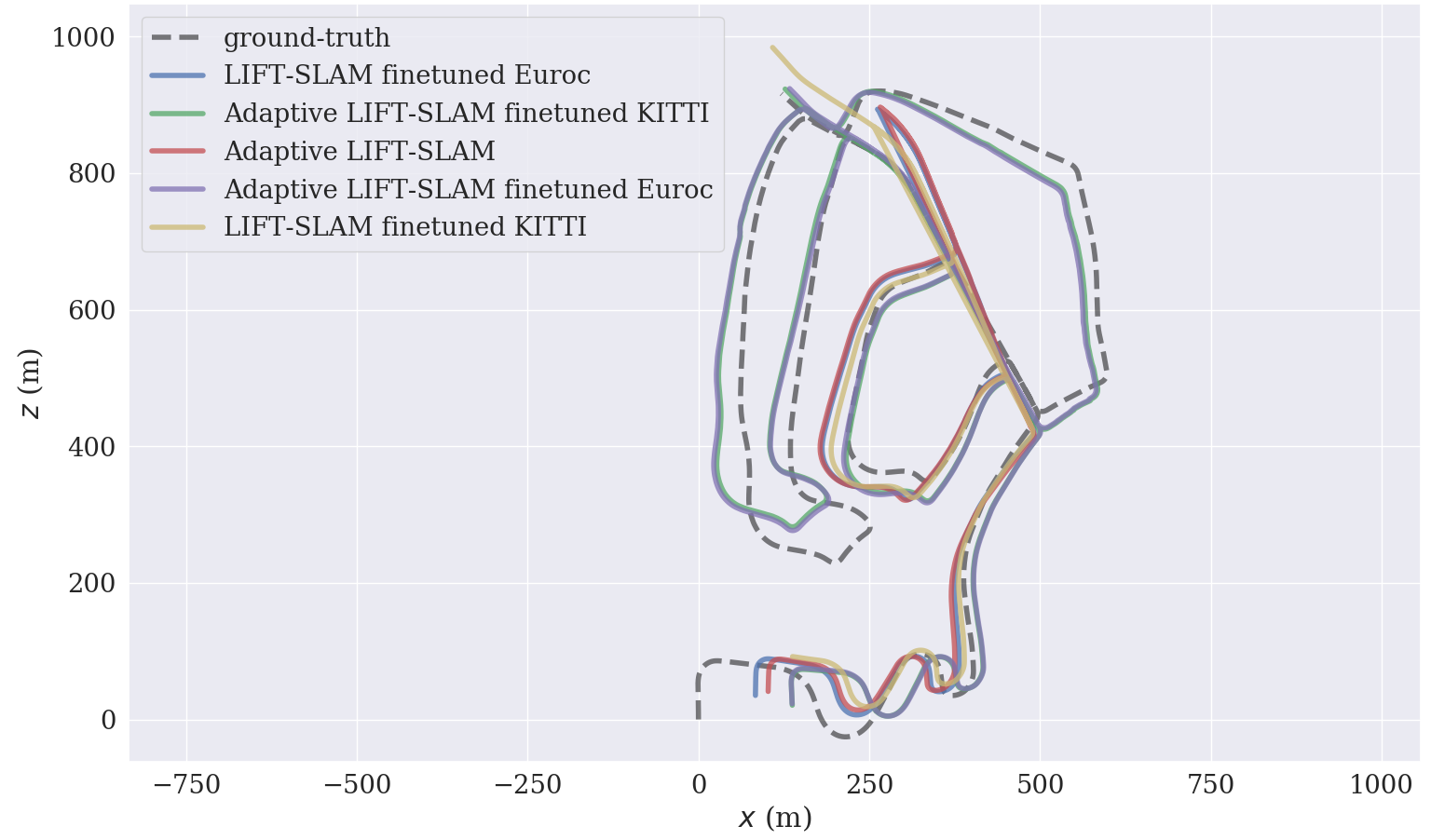}
\label{fig:kitti-02}}

\subfloat[][KITTI 03]{
\includegraphics[width=0.44\textwidth, height=4cm]{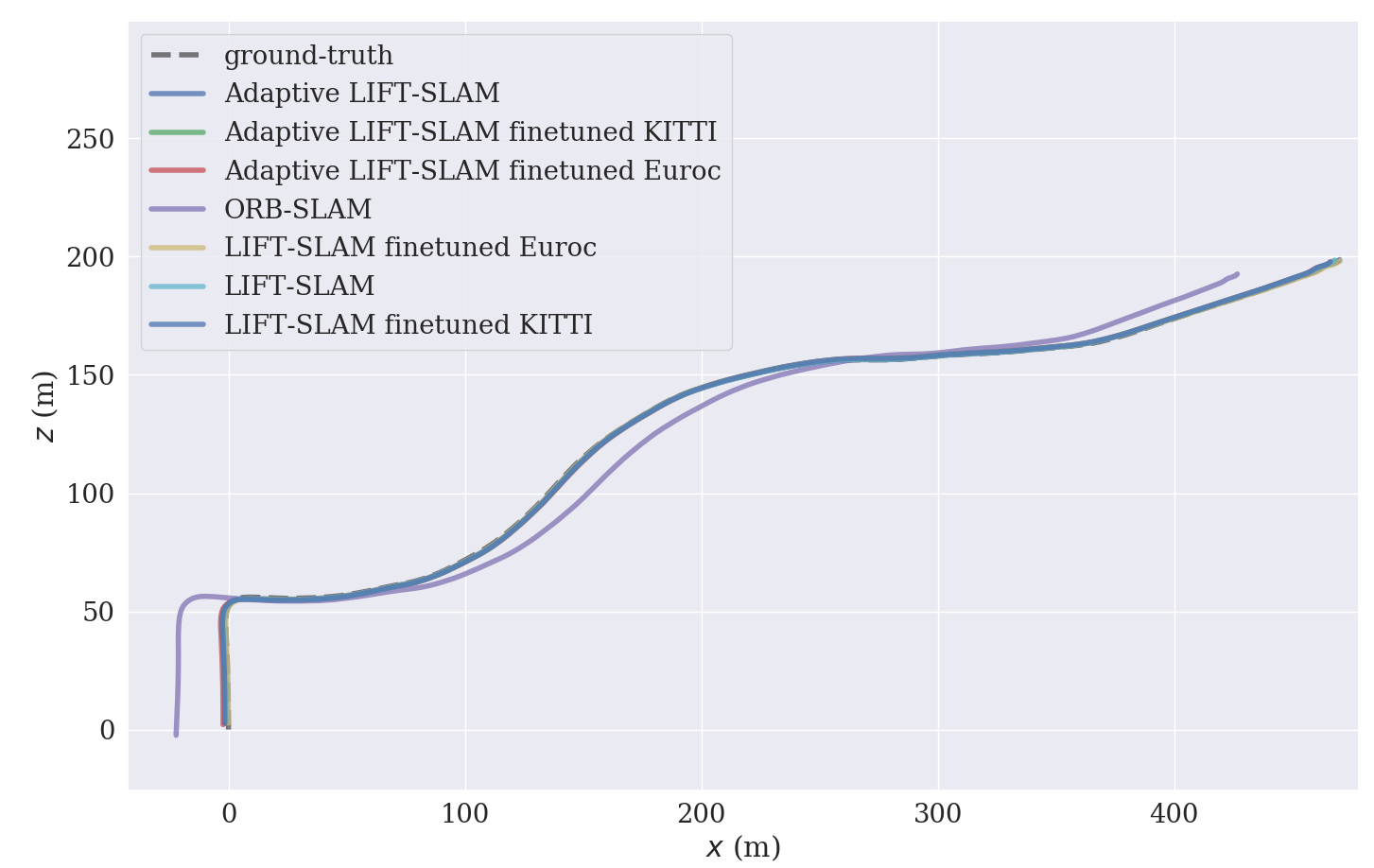}
\label{fig:kitti-03}}
\qquad
\subfloat[][KITTI 05]{
\includegraphics[width=0.44\textwidth, height=4cm]{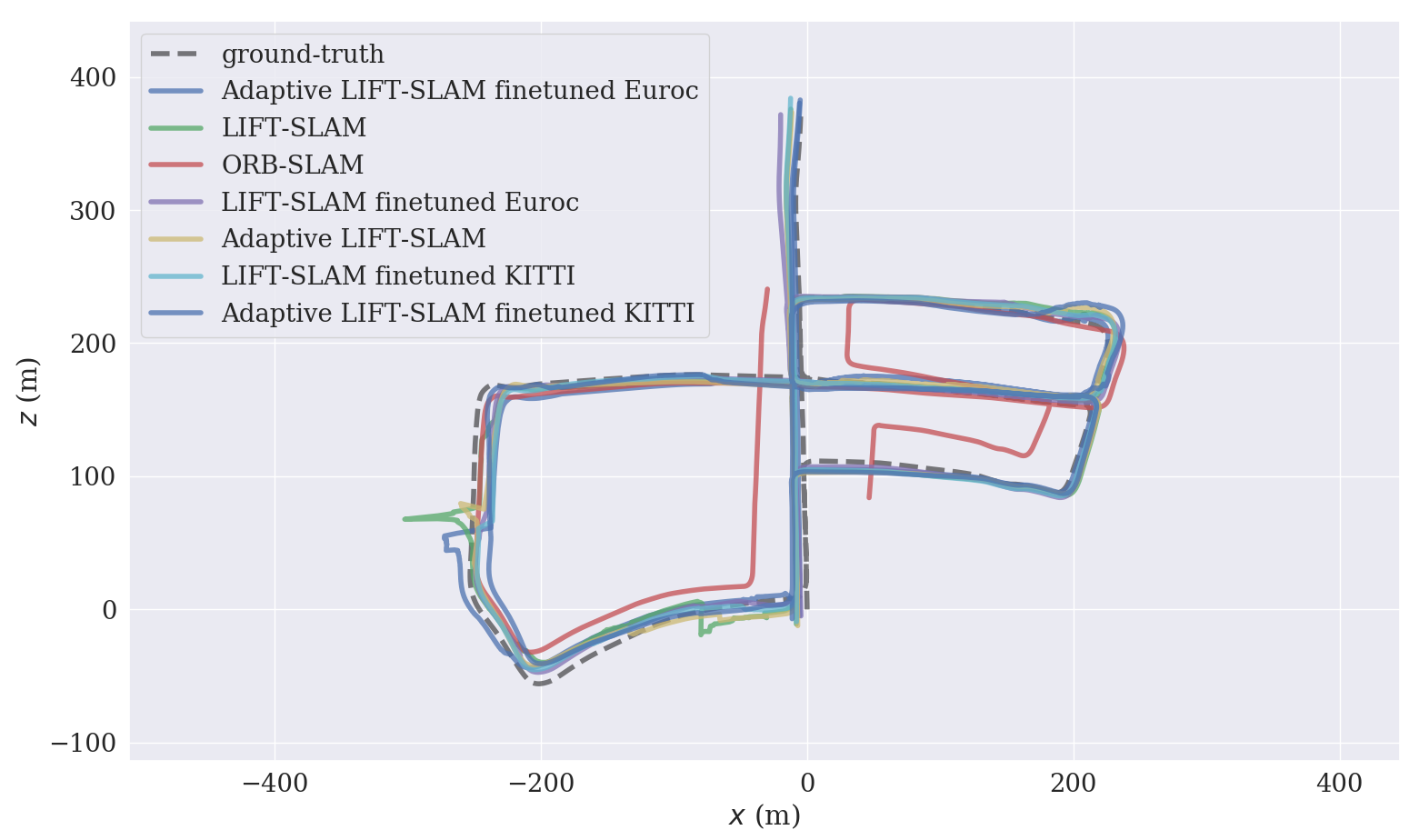}
\label{fig:kitti-05}}

\subfloat[][KITTI 06]{
\includegraphics[width=0.44\textwidth, height=4cm]{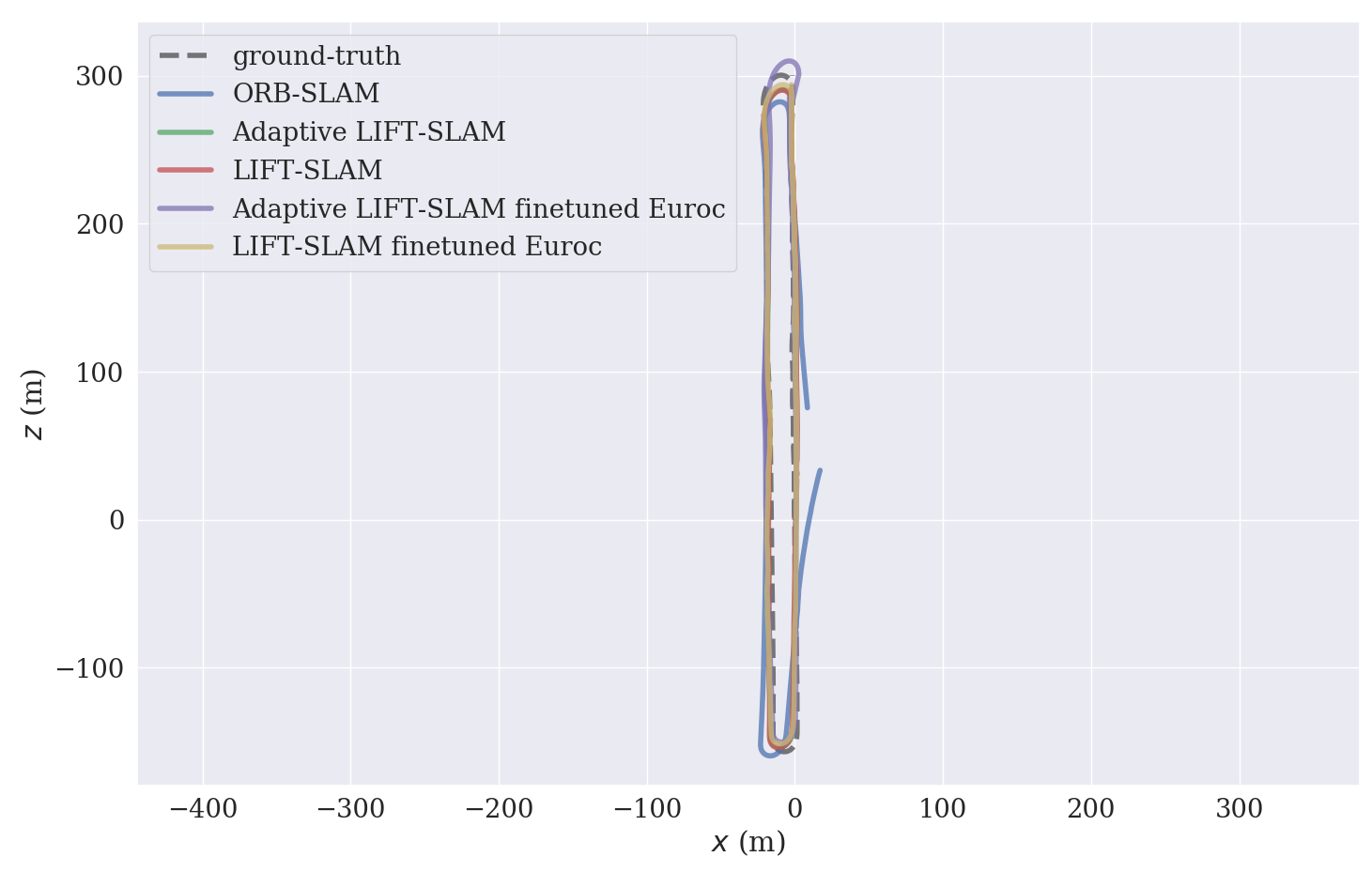}
\label{fig:kitti-06}}
\qquad
\subfloat[][KITTI 07]{
\includegraphics[width=0.44\textwidth, height=4cm]{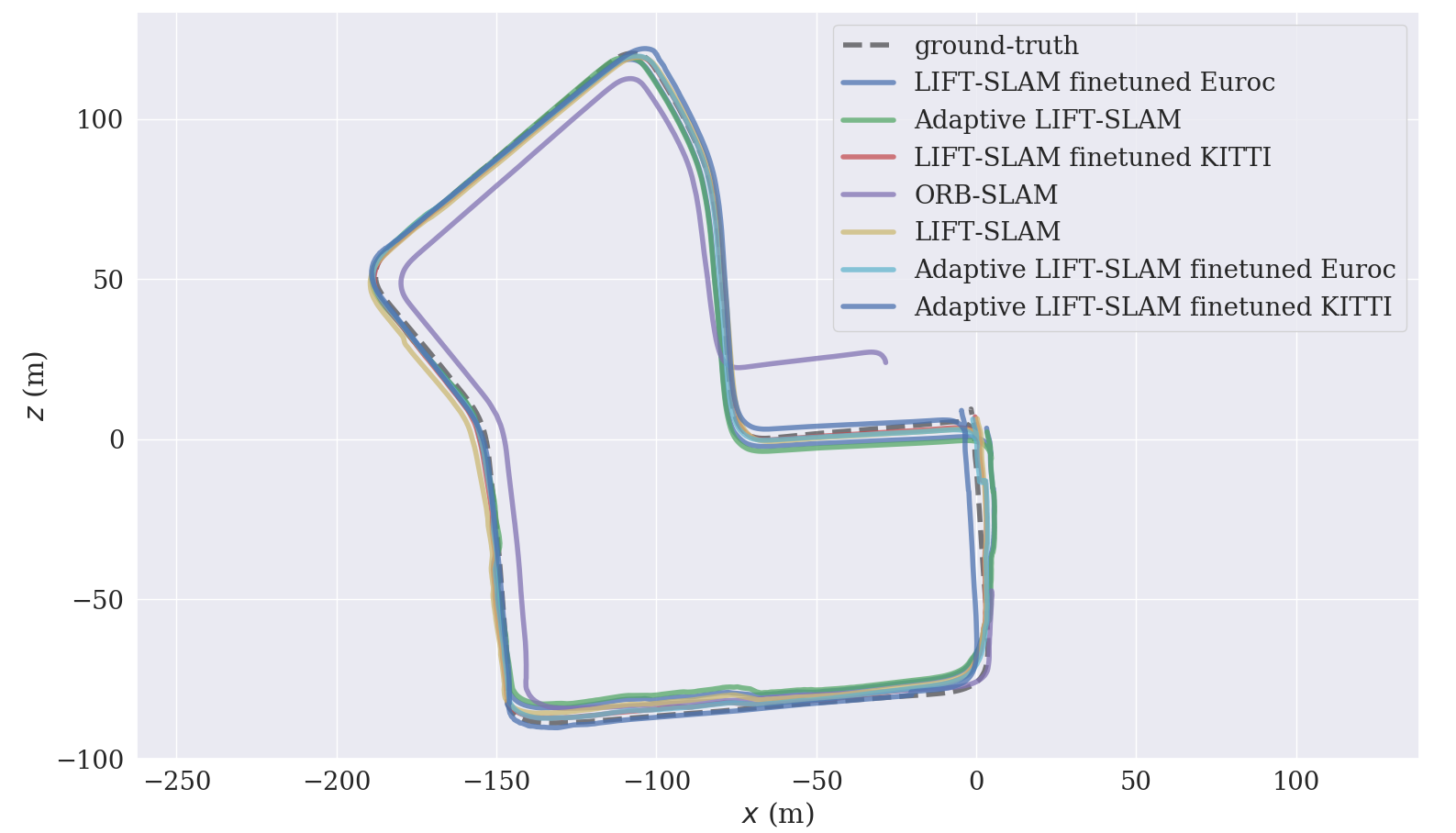}
\label{fig:kitti-07}}

\subfloat[][KITTI 08]{
\includegraphics[width=0.44\textwidth, height=4cm]{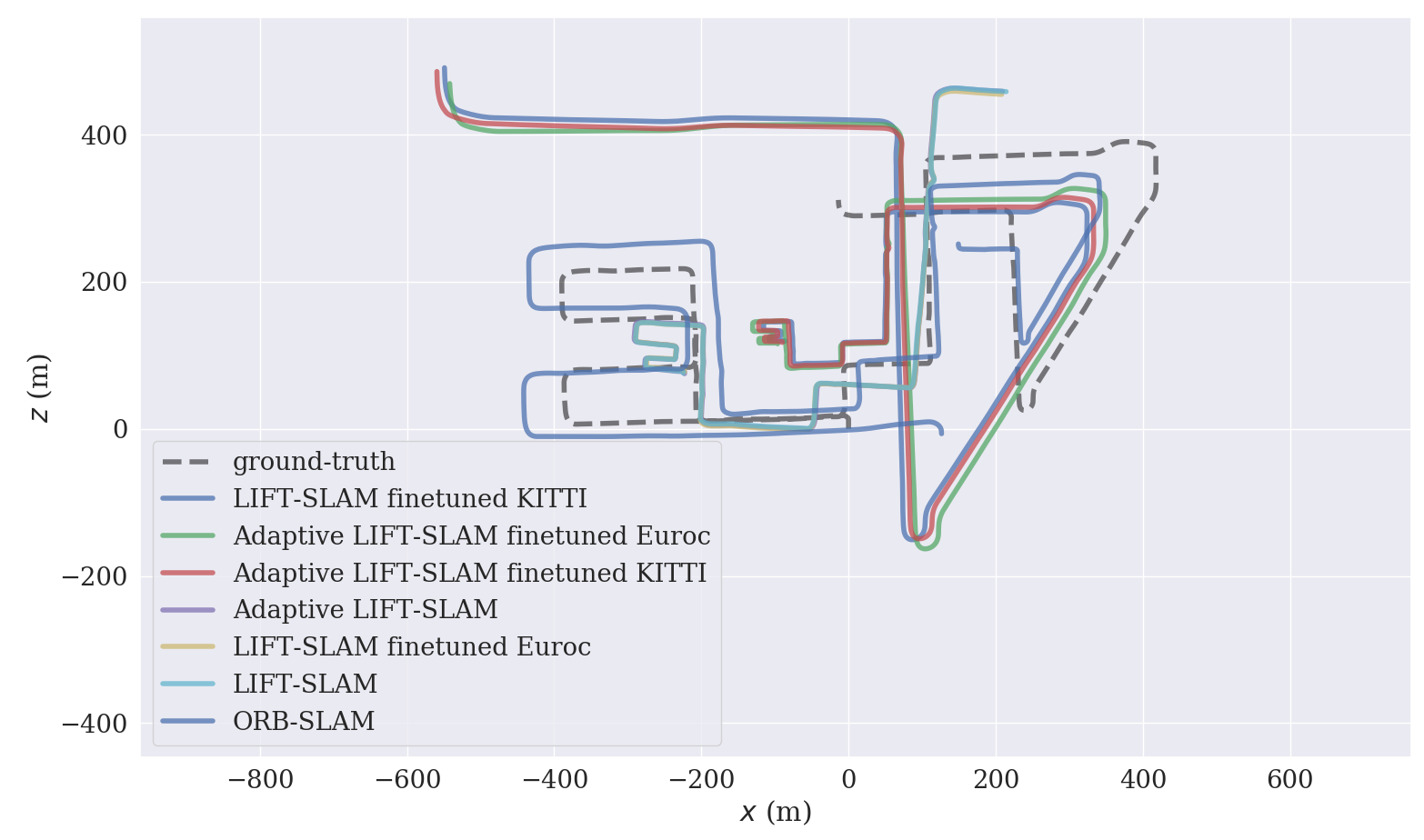}
\label{fig:kitti-08}}
\qquad
\subfloat[][KITTI 10]{
\includegraphics[width=0.44\textwidth, height=4cm]{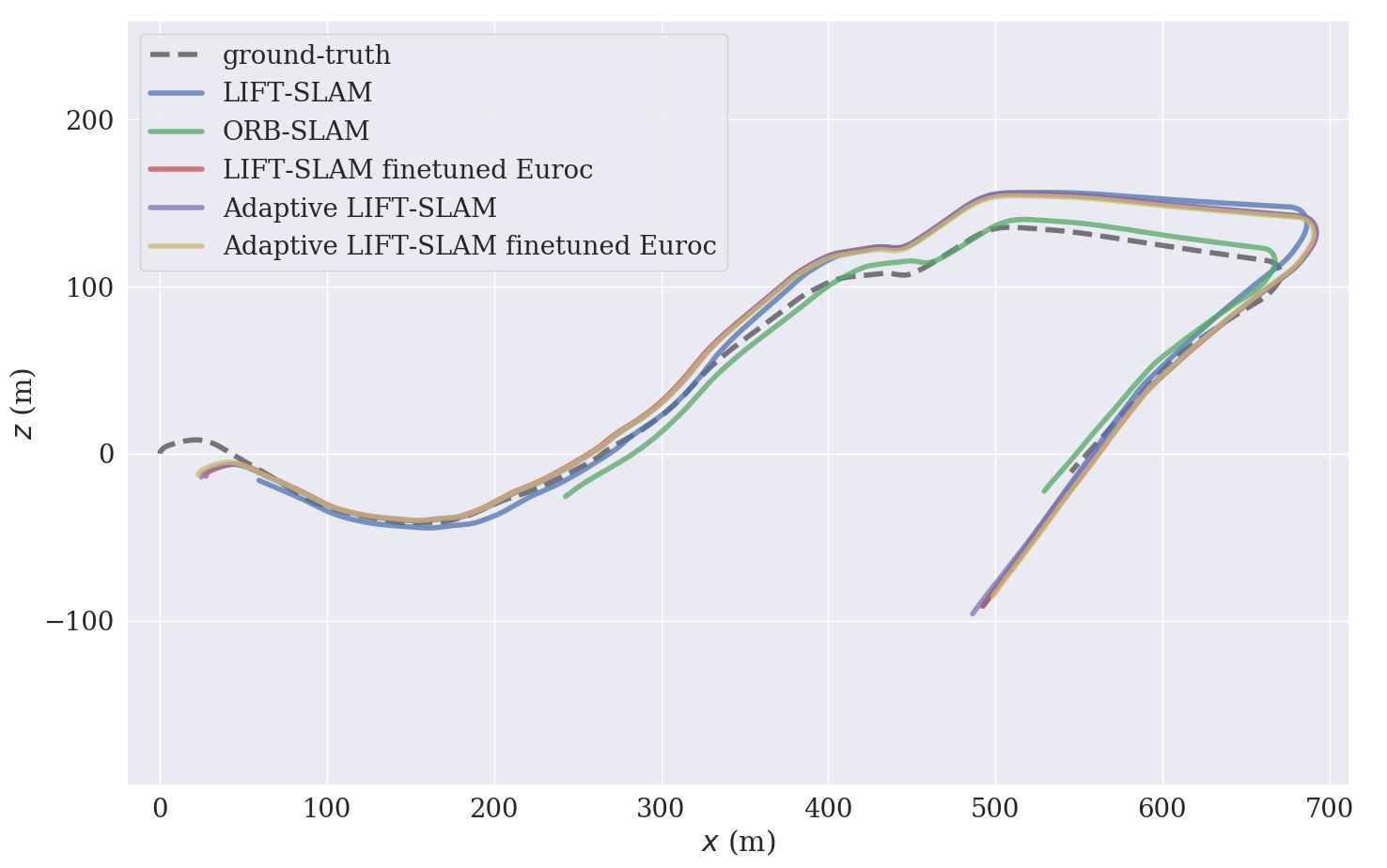}
\label{fig:kitti-10}}

\caption{Results in KITTI dataset.}
\label{fig:all-traj-kitti}
\end{figure}

Figure \ref{fig:all-traj-kitti} shows the qualitative comparison between the algorithms in the KITTI dataset. In sequences $00$ (Fig. \ref{fig:kitti-00}) and $02$ (Fig. \ref{fig:kitti-02}) most of the algorithms could not track the entire trajectory, except for Adaptive LIFT-SLAM fine-tuned with Euroc. Figure \ref{fig:kitti-03} shows the difference in performance in smaller sequences between ORB-SLAM and all LIFT-SLAM versions. Moreover, in sequences $05$, $06$, and $07$, ORB-SLAM could not detect loop-closure, thus, its trajectories has an accumulated drift as shown in Figures \ref{fig:kitti-05}, \ref{fig:kitti-06} and \ref{fig:kitti-07}. On the other hand, in Figure \ref{fig:kitti-08}, we can notice that none of the algorithms could detect loop closure in the sequence of $08$. Therefore, all estimated trajectories have a big error accumulated over the entire sequence.

Table \ref{tab:all-results-euroc} shows the quantitative comparison between the algorithms in Euroc dataset. We can notice that ORB-SLAM has the smallest average in 3 sequences. Moreover, the proposed versions of LIFT-SLAM performed better than original LIFT-SLAM in all sequences. The algorithm's performance improved even when we used the KITTI dataset to fine-tune the LIFT network. Therefore we confirmed that this network version also learned essential features from the dataset.

\begin{table}[!h]
\centering
\resizebox{0.8\textwidth}{!}{\begin{tabular}{|c|cccccc|}
\hline \textbf{Algorithm} & \textbf{MH\_01} & \textbf{MH\_02} & \textbf{MH\_03} & \textbf{MH\_04} & \textbf{V1\_01} & \textbf{V1\_03} \\ \hline
ORB-SLAM & 0.048 & 0.037 & \textbf{0.040} & 0.432 & \textbf{0.100} & \textbf{0.370} \\
LIFT-SLAM & 0.062 & 0.227 & 0.144 & 1.859 & X & X \\
LIFT-SLAM fine-tuned with KITTI & 0.115 & 0.042 & 0.055 & \textbf{0.117} & 0.117 & X \\
LIFT-SLAM fine-tuned with Euroc & 0.117 & 0.062 & 0.053 & - & 0.150 & - \\
Adaptive LIFT-SLAM & 0.046 & \textbf{0.034} & X & X & 0.101 & X \\
Adaptive LIFT-SLAM fine-tuned with KITTI & 0.455 & X & 0.116 & X & 0.194 & X \\
Adaptive LIFT-SLAM fine-tuned with Euroc & \textbf{0.044} & 0.053 & 0.049 & - & 0.157 & - \\ \hline
\end{tabular}}

\caption{ATE (m) comparison between ORB-SLAM and all versions of LIFT-SLAM in the Euroc dataset. We fill with "X" the sequences unavailable due to tracking failure, and with "-" sequences, we do not execute the algorithm to avoid biased results. We highlight the smaller average in each metric.}
\label{tab:all-results-euroc}
\end{table}

\begin{figure}
\centering
\subfloat[][Euroc MH\_01]{
\includegraphics[width=0.44\textwidth, height=4cm]{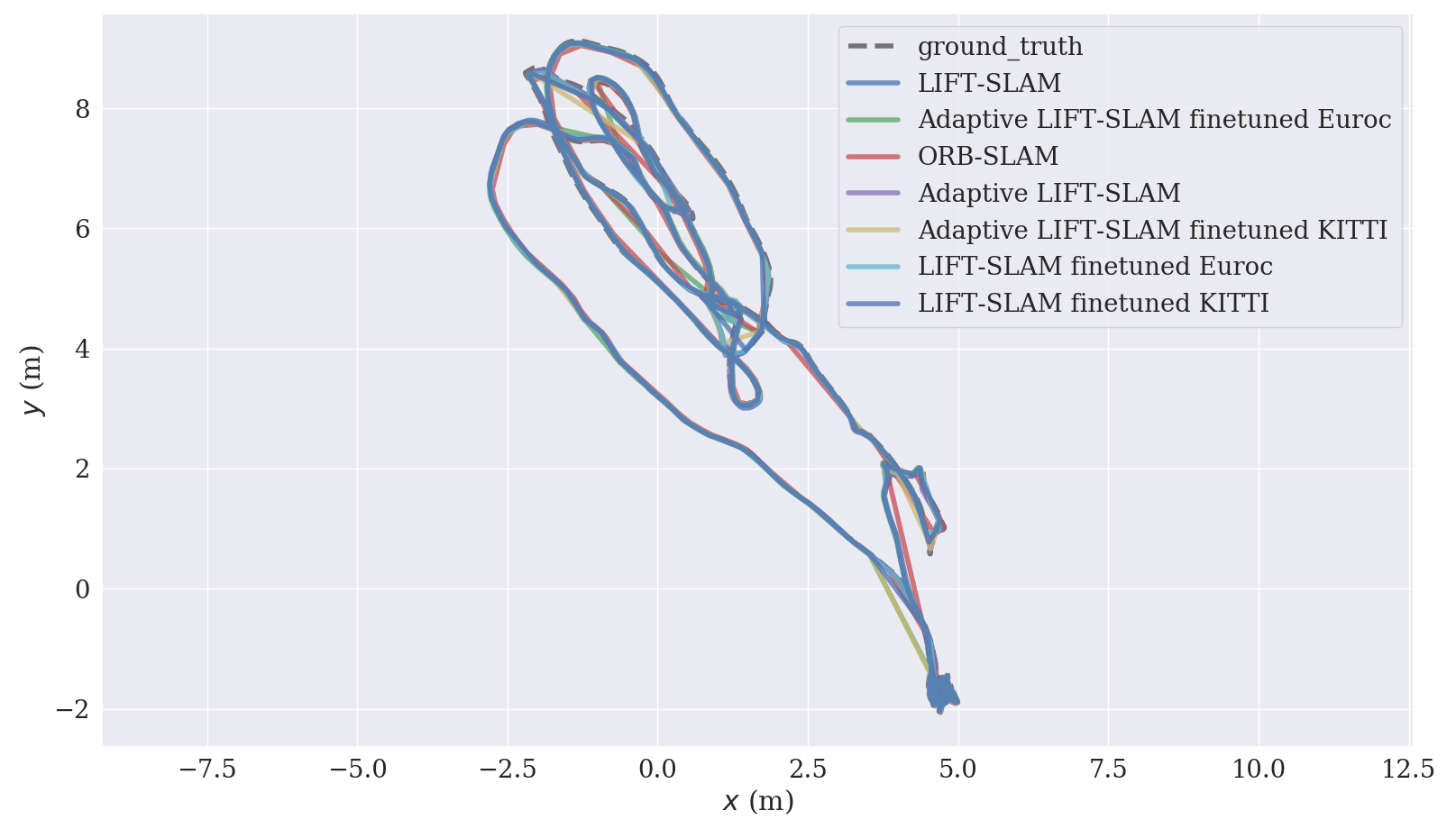}
\label{fig:mh01}}
\qquad
\subfloat[][Euroc MH\_02]{
\includegraphics[width=0.44\textwidth, height=4cm]{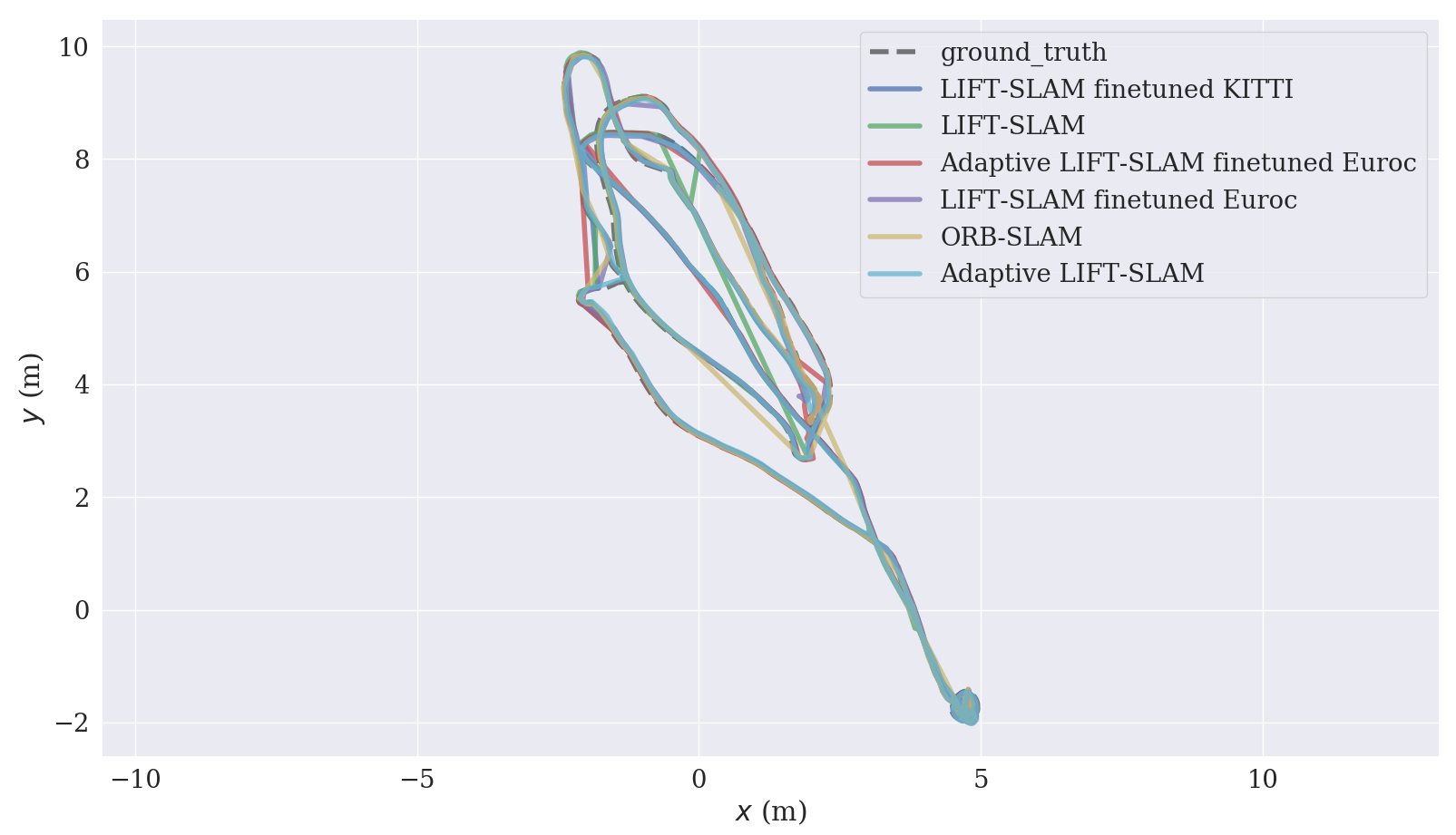}
\label{fig:mh02}}

\subfloat[][Euroc MH\_03]{
\includegraphics[width=0.44\textwidth, height=4cm]{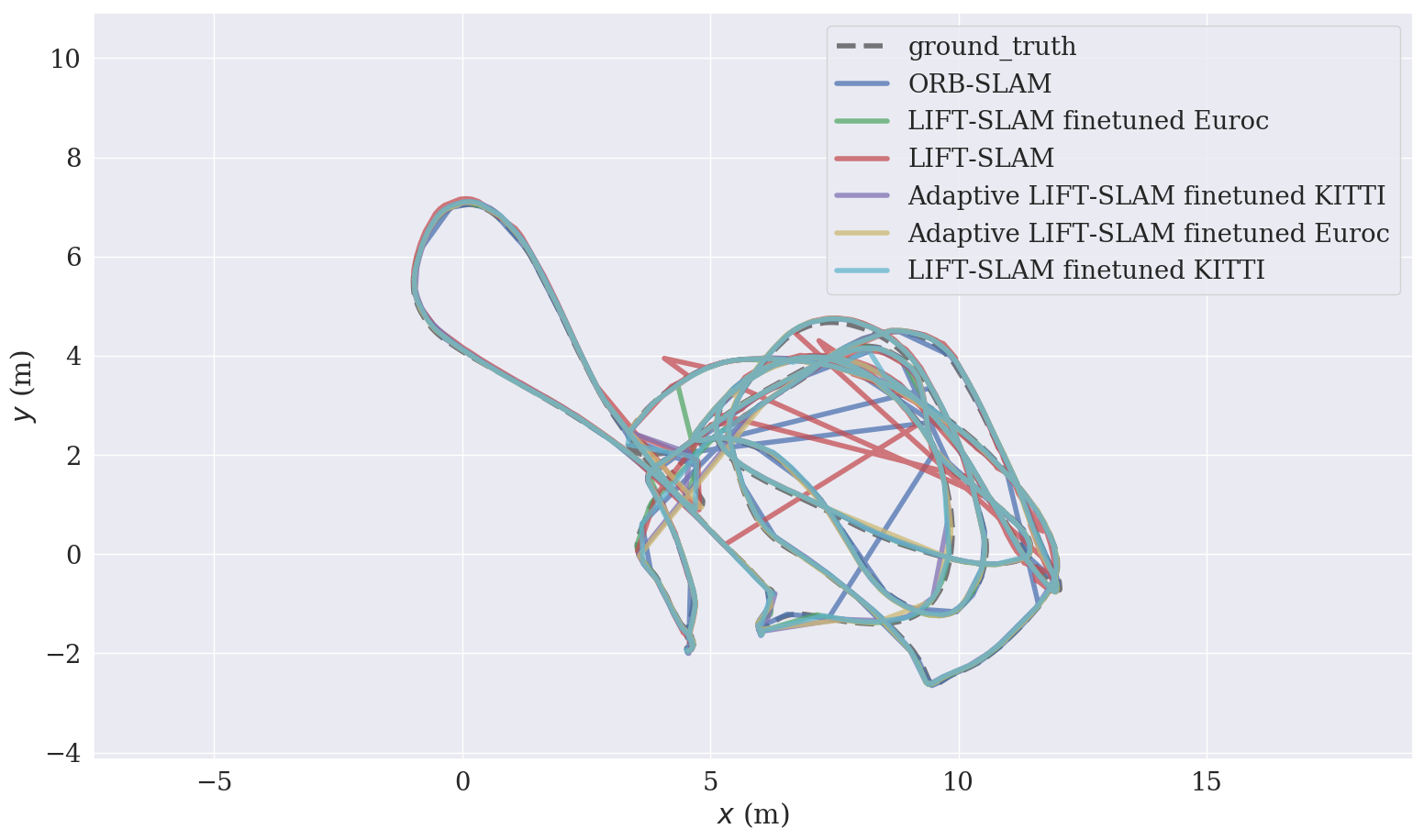}
\label{fig:mh03}}
\qquad
\subfloat[][Euroc MH\_04]{
\includegraphics[width=0.44\textwidth, height=4cm]{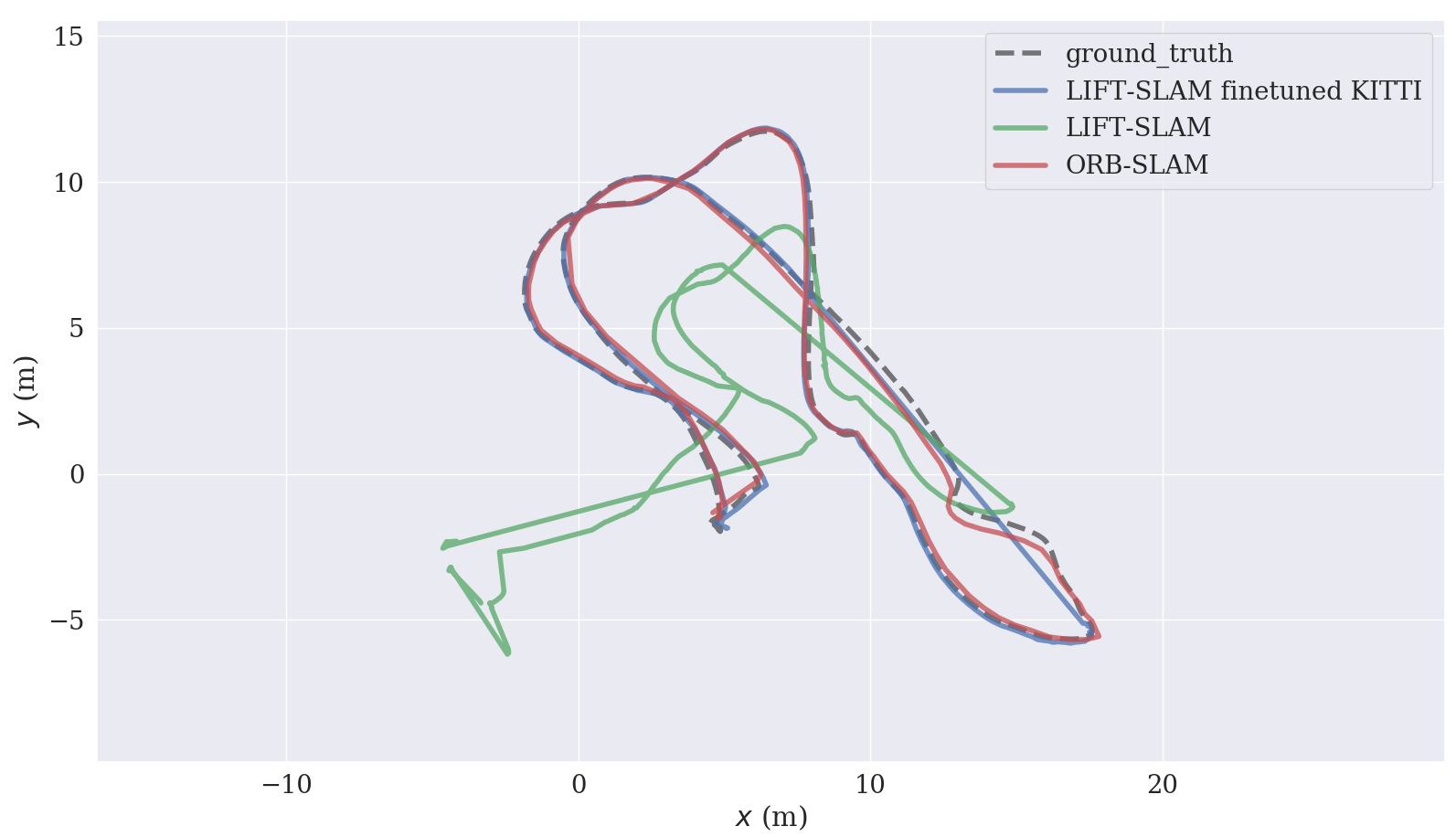}
\label{fig:mh04}}

\subfloat[][Euroc V1\_01]{
\includegraphics[width=0.44\textwidth, height=4cm]{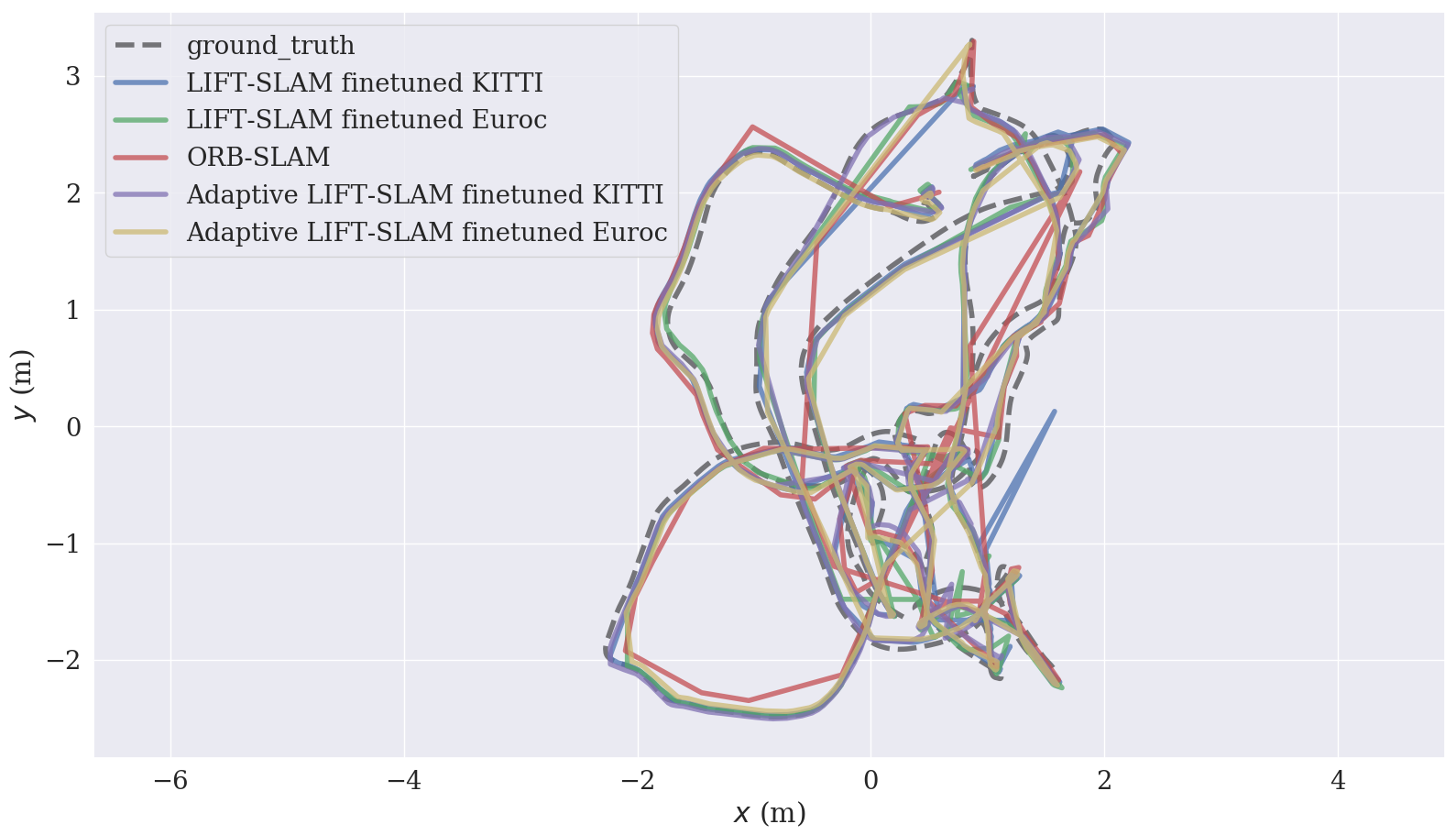}
\label{fig:v101}}

\caption{Results in Euroc dataset.}
\label{fig:all-traj-euroc}
\end{figure}

Figure \ref{fig:all-traj-euroc} shows the qualitative comparison between the algorithms in Euroc dataset. 
In sequence MH\_01 (Fig. \ref{fig:mh01}), all algorithms had a good performance. Moreover, in MH\_02 (Fig. \ref{fig:mh02}), most of the algorithms also performed well, except for Adaptive LIFT-SLAM finetuned with KITTI that failed to compute the pose. Figure \ref{fig:mh03} shows that the only algorithm that could track the entire trajectory was Adaptive LIFT-SLAM finetuned with Euroc. In sequence MH\_04 (Fig. \ref{fig:mh04}), LIFT-SLAM had a terrible performance, but its version finetuned with KITTI is more similar to the ground-truth, which shows that finetuning the network was effective. Lastly, in sequence V1\_01 (Fig. \ref{fig:v101}), all LIFT-SLAM versions presented in the Figure could track the trajectory while ORB-SLAM lost track multiple times. Therefore, considering the quantitative and qualitative results of all versions of LIFT-SLAM, Adaptive LIFT-SLAM finetuned with Euroc sequences is the one with better overall results.

\subsection{Robustness tests}

\begin{figure}
\centering
\subfloat[][$\gamma < 1$.]{
\includegraphics[width=0.6\textwidth]{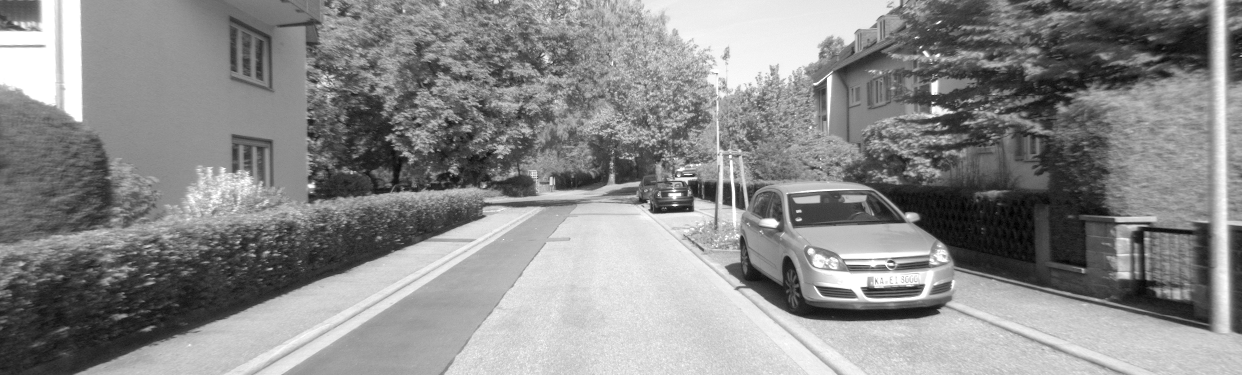}
\label{fig:gamma0.25}}

\subfloat[][$\gamma > 1$.]{
\includegraphics[width=0.6\textwidth]{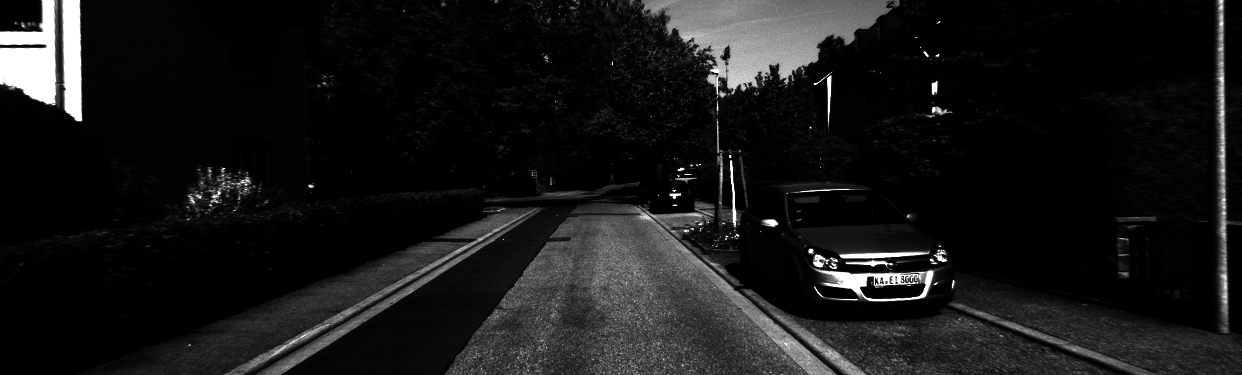}
\label{fig:gamma4}}

\subfloat[][Truncation in $Q_1$.]{
\includegraphics[width=0.6\textwidth]{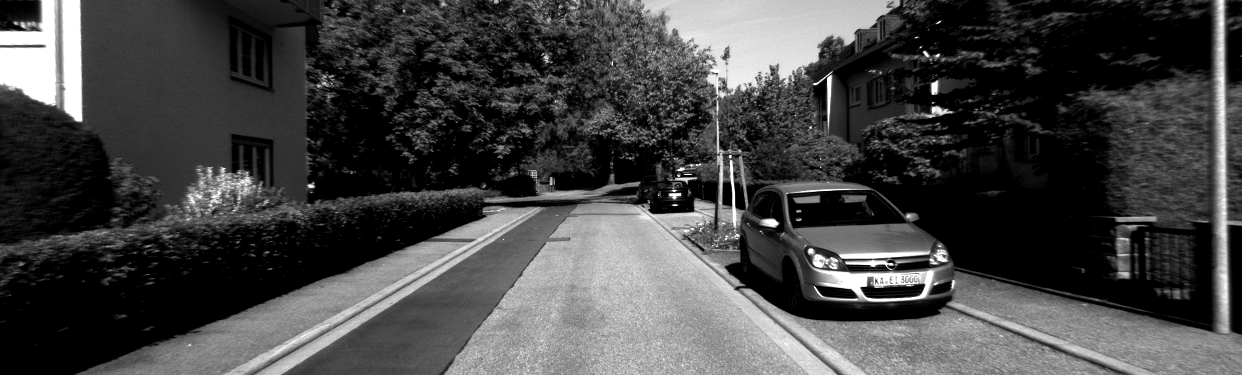}
\label{fig:quantile1}}

\subfloat[][Truncation in $Q_3$.]{
\includegraphics[width=0.6\textwidth]{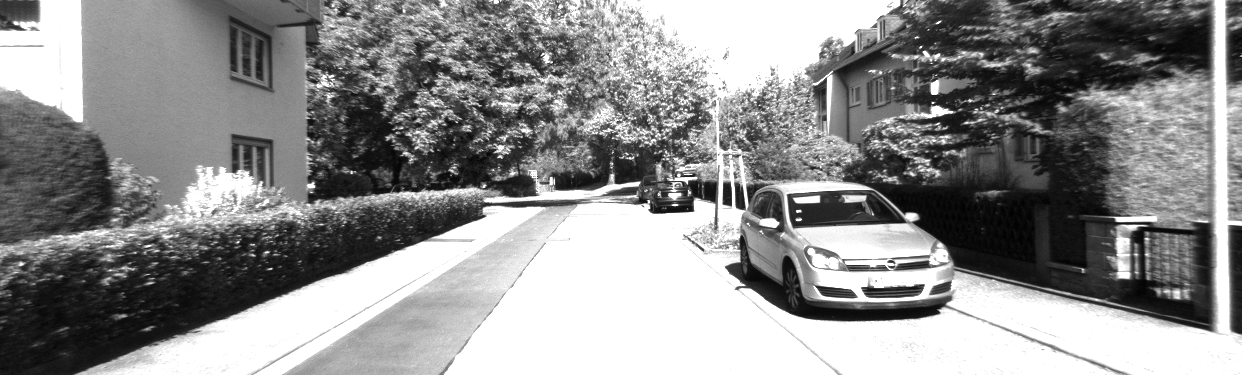}
\label{fig:quantile3}}

\caption{Examples of the distortions applied to images to test the robustness of the algorithms.}
\label{fig:distortions}
\end{figure}

To test our system's robustness to camera sensor noise, we created different image distortion in some sequences of KITTI and Euroc simulating camera ill exposure conditions. These scenarios were emulated with the application of gamma power transformation and quantile-based truncation, as proposed in \cite{emulate-exposure}. More details about these operations are described next:

\begin{itemize}
    \item Gamma power transformation: This transformation creates a new image $I' $ from image $I$ by applying: $I' =I^{\gamma}$. We used four values of $\gamma$: $0.25$, $0.5$, $2$ and $4$. Values of $\gamma < 1$ results in data loss for bright regions emulating camera overexposing, as shown in Figure \ref{fig:gamma0.25}. $\gamma > 1$ results data loss for dark regions emulating camera underexposing \cite{emulate-exposure} (Fig. \ref{fig:gamma4});
    
    \item Quantile-based truncation: We have truncated the first ($Q_1$) and third ($Q_3$) quantiles of the pixels' intensities distribution to reproduce the effects of low dynamic range imaging sensors. When truncating pixels in $Q_1$ we emulate sensor underexposing (Fig. \ref{fig:quantile1}), and in $Q_3$ we emulate sensor overexposing (Fig. \ref{fig:quantile3}).
\end{itemize}

We also tried to apply a salt-and-pepper noise in the sequences to simulate the malfunctioning of the camera's sensor cell \cite{emulate-exposure}. However, in this scenario, none of the algorithms were capable of initializing the map. Therefore, we do not present results for this case.

As Adaptive LIFT-SLAM fine-tuned with Euroc sequences obtained the best overall results, we tested this algorithm under the described scenarios and compared its performance with ORB-SLAM under the same scenarios. The quantitative results of the tests are shown in table \ref{tab:distortion-results}. ORB-SLAM could not track the camera's pose with some distortion in sequences KITTI 03, 06, and 10, while LIFT-SLAM failed in some cases for sequences KITTI 10 and Euroc MH\_02. 
We can also notice that in most of the sequences, LIFT-SLAM improved its performance when we applied the distortions. This fact occurs because the distortions remove some outliers from the images, which allows the algorithms to select better keypoints.

\begin{table}[!h]
\centering
\resizebox{\textwidth}{!}{\begin{tabular}{|c|c|ccc|ccc|}
\hline
 Sequence & \textbf{Distortion} &  \multicolumn{3}{|c|}{\textbf{ORB-SLAM}} &  \multicolumn{3}{c|}{\textbf{LIFT-SLAM}} \\ \hline
 &  & $\mathbf{RPE_{trans}}$ \textbf{(\%)} & $\mathbf{RPE_{rot}}$ \textbf{(deg/m)} & \textbf{ATE (m)} & $\mathbf{RPE_{trans}}$ \textbf{(\%)} & $\mathbf{RPE_{rot}}$ \textbf{(deg/m)} & \textbf{ATE (m)} \\ \hline
 & no distortion & 9.75 & 2.78 &15.13 & 1.46 & 0.34 & 2.23 \\
 & $\gamma = 0.25$ & 7.68 & 1.95 & 11.72 & 1.02 & 0.40 & 1.23  \\
 & $\gamma = 0.5$ & 8.25 & 2.24 & 11.38 & 1.28 & 0.36 & 1.74 \\
KITTI 03 & $\gamma = 2$ & X & X & X & 1.07 & 0.51 & 1.47 \\
& $\gamma = 4$ & X & X & X & 2.82 & 0.70 & 5.23 \\
& Truncation in $Q_1$ & 8.34 & 1.41 & 13.63 & 1.36 & 0.45 & 2.07  \\
& Truncation in $Q_3$ & 9.78 & 2.23 & 15.66 & 1.10 & 0.46 & 1.27 \\ \hline

& no distortion & 8.11 & 2.88 & 20.26 & 12.24 & 2.91 & 30.38 \\
& $\gamma = 0.25$ & 8.97 & 2.11 & 21.85 & 7.94 & 2.22 & 19.07 \\
& $\gamma = 0.5$ & 9.55 & 2.16 & 24.11 & 7.61 & 2.27 & 18.19 \\
KITTI 06 & $\gamma = 2$ & 11.69 & 4.50 & 27.24 & 6.91 & 2.30 & 16.22  \\
& $\gamma = 4$ & 11.12 & 5.71 & 28.10 & 8.29 & 2.68 & 18.60  \\
& Truncation in $Q_1$ & 15.01 & 4.69 & 36.52 & 6.44 & 2.36 & 15.09  \\
& Truncation in $Q_3$ & X & X & X & 8.09 & 2.29 & 19.33 \\ \hline

& no distortion & 7.43 & 3.58 & 13.47 & 2.42 & 4.02 & 3.63 \\
& $\gamma = 0.25$ & 7.61 & 2.42 & 12.54 & 2.09 & 3.17 & 3.30\\
& $\gamma = 0.5$ & 6.37 & 2.02 & 9.58 & 1.99 & 3.88 & 2.86 \\
KITTI 07 & $\gamma = 2$ & 5.89 & 2.11 & 9.36 & 3.43 & 4.32 & 5.91 \\
& $\gamma = 4$ & 6.61 & 7.06 & 7.84 & 8.03 & 7.40 & 16.13\\
& Truncation in $Q_1$ & 8.50 & 2.47 & 10.69 & 2.45 & 4.10 & 3.58 \\
& Truncation in $Q_3$ & 7.01 & 2.40 & 7.08 & 2.69 & 3.77 & 4.49 \\ \hline

& no distortion & 8.65 & 3.62 & 19.94 & 9.72 & 2.24 & 29.87 \\
& $\gamma = 0.25$ & 13.52 & 3.01 & 26.56 & 10.72 & 2.15 & 30.77 \\
& $\gamma = 0.5$ & 12.40 & 3.95 & 25.55 & 10.03 & 2.24 & 31.93 \\
KITTI 10 & $\gamma = 2$ & 16.88 & 4.59 & 28.07 & X & X & X\\
& $\gamma = 4$ & X & X & X & X & X & X \\
& Truncation in $Q_1$ & 20.79 & 2.95 & 34.79 & X & X & X \\
& Truncation in $Q_3$ & X & X & X & X & X & X\\ \hline

& no distortion & - & - & 0.037 & - & - & 0.053\\
& $\gamma = 0.25$ & - & - & 0.055 & - & - & 0.035 \\
& $\gamma = 0.5$ & - & - & 0.040 & - & - & 0.039 \\
Euroc MH\_02 & $\gamma = 2$ & - & - & 0.061 & - & - & 0.037 \\
& $\gamma = 4$ & - & - & 0.010 & - & - & 0.194 \\
& Truncation in $Q_1$ & - & - & 0.039 & - & - & 0.043 \\
& Truncation in $Q_3$ & - & - & 0.043 & - & - & X \\ \hline

\end{tabular}}
\caption[]{Results of the robustness tests. The LIFT-SLAM version used in these tests is the adaptive fine-tuned with Euroc sequences. We fill with "X" the sequences unavailable due to tracking failure and with "-" the sequences we do not execute the algorithms.}
\label{tab:distortion-results}
\end{table}

\begin{figure}
\centering
\subfloat[ORB-SLAM in KITTI 03]{
\includegraphics[width=0.4\textwidth, height=3cm]{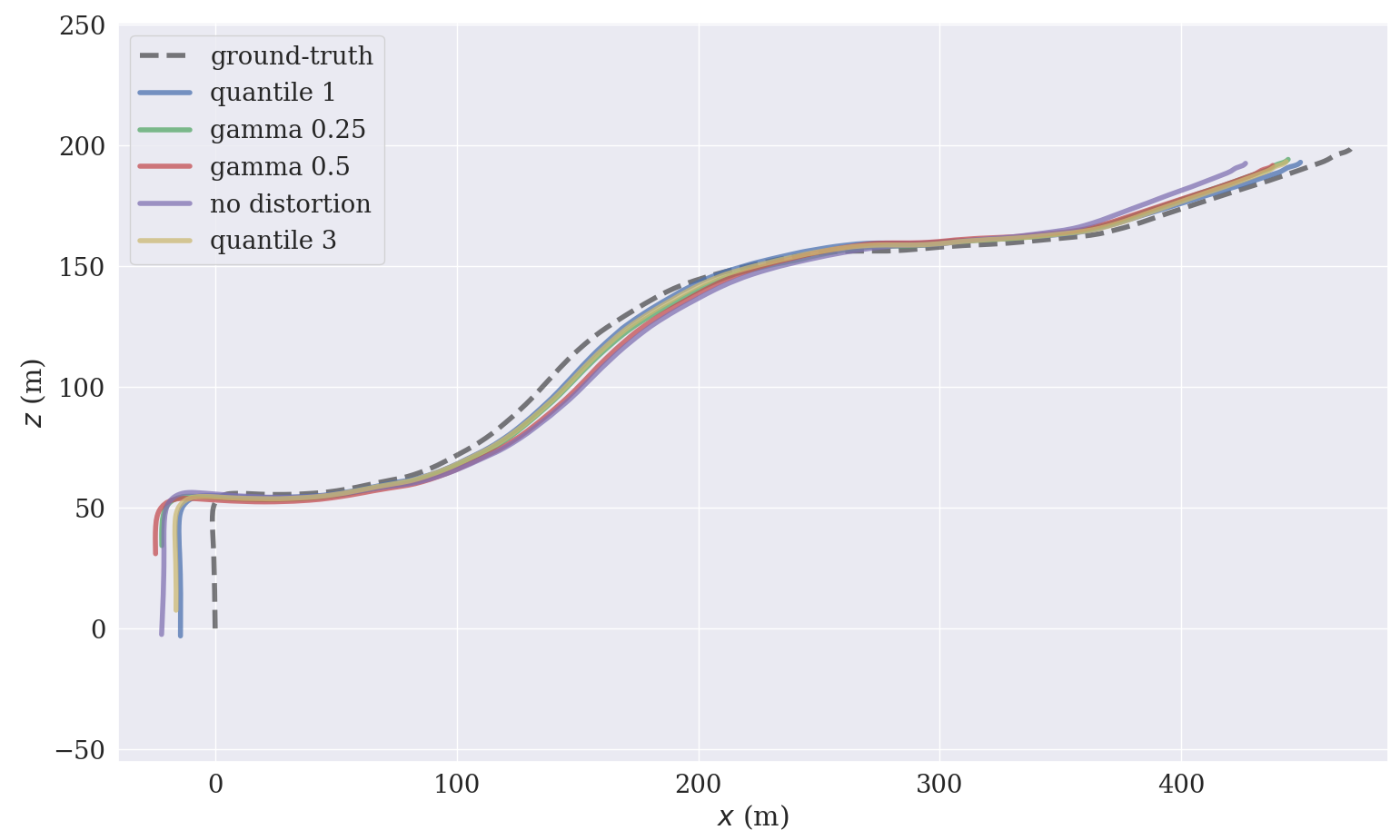}
\label{fig:orbslam-noise-03}}
\qquad
\subfloat[LIFT-SLAM in KITTI 03]{
\includegraphics[width=0.4\textwidth, height=3cm]{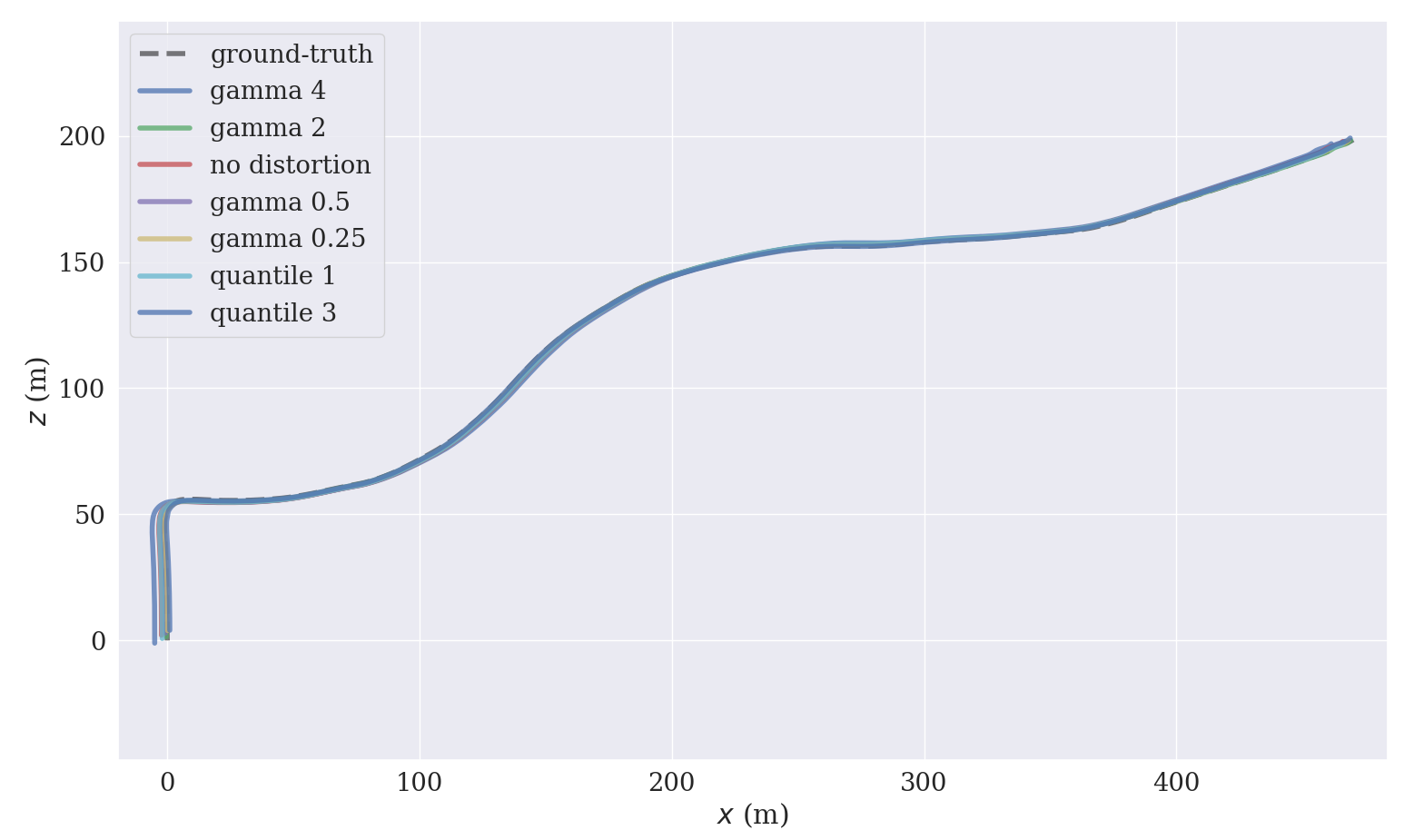}
\label{fig:liftslam-noise-03}}

\subfloat[ORB-SLAM in KITTI 06]{
\includegraphics[width=0.4\textwidth, height=3cm]{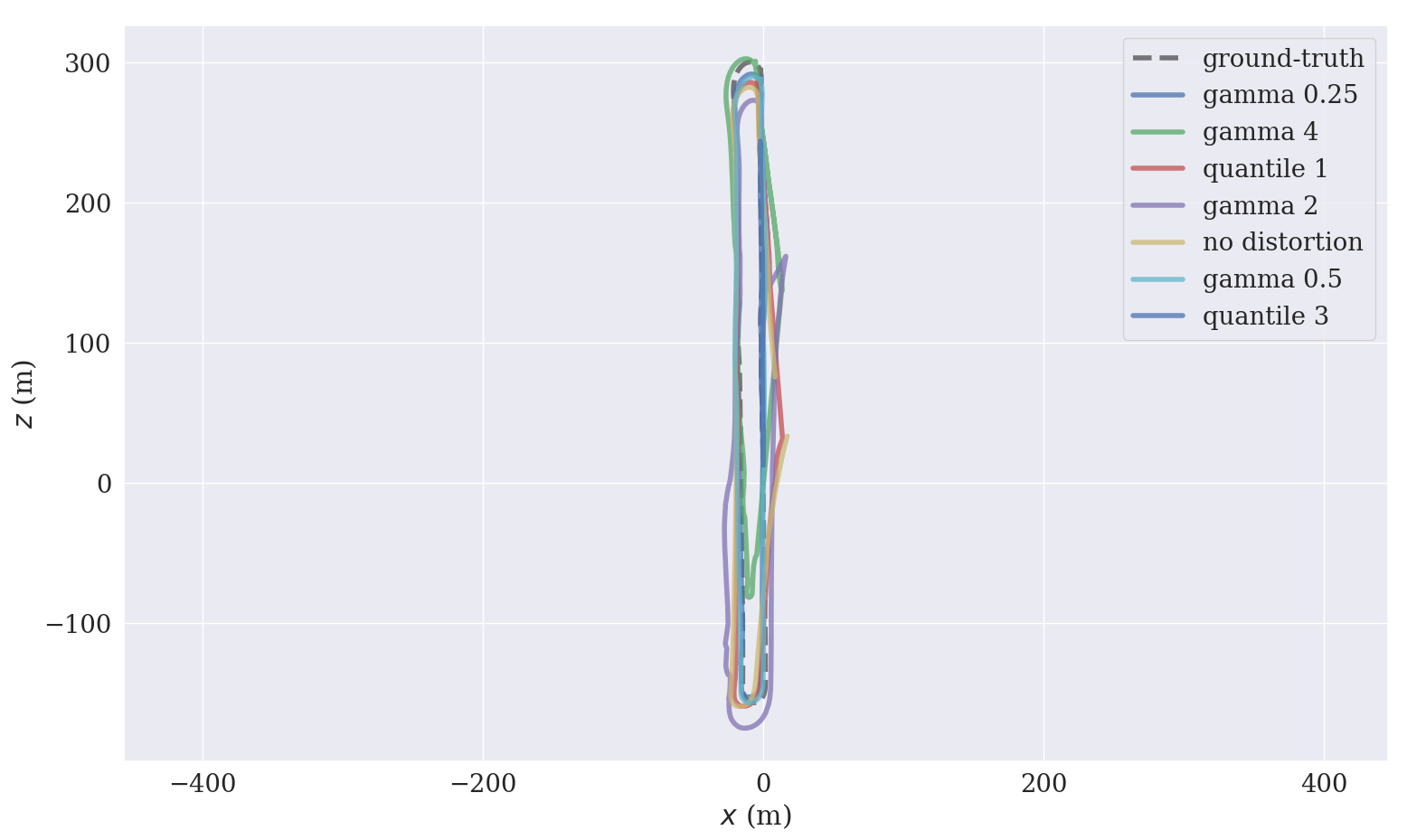}
\label{fig:orbslam-noise-06}}
\qquad
\subfloat[LIFT-SLAM in KITTI 06]{
\includegraphics[width=0.4\textwidth, height=3cm]{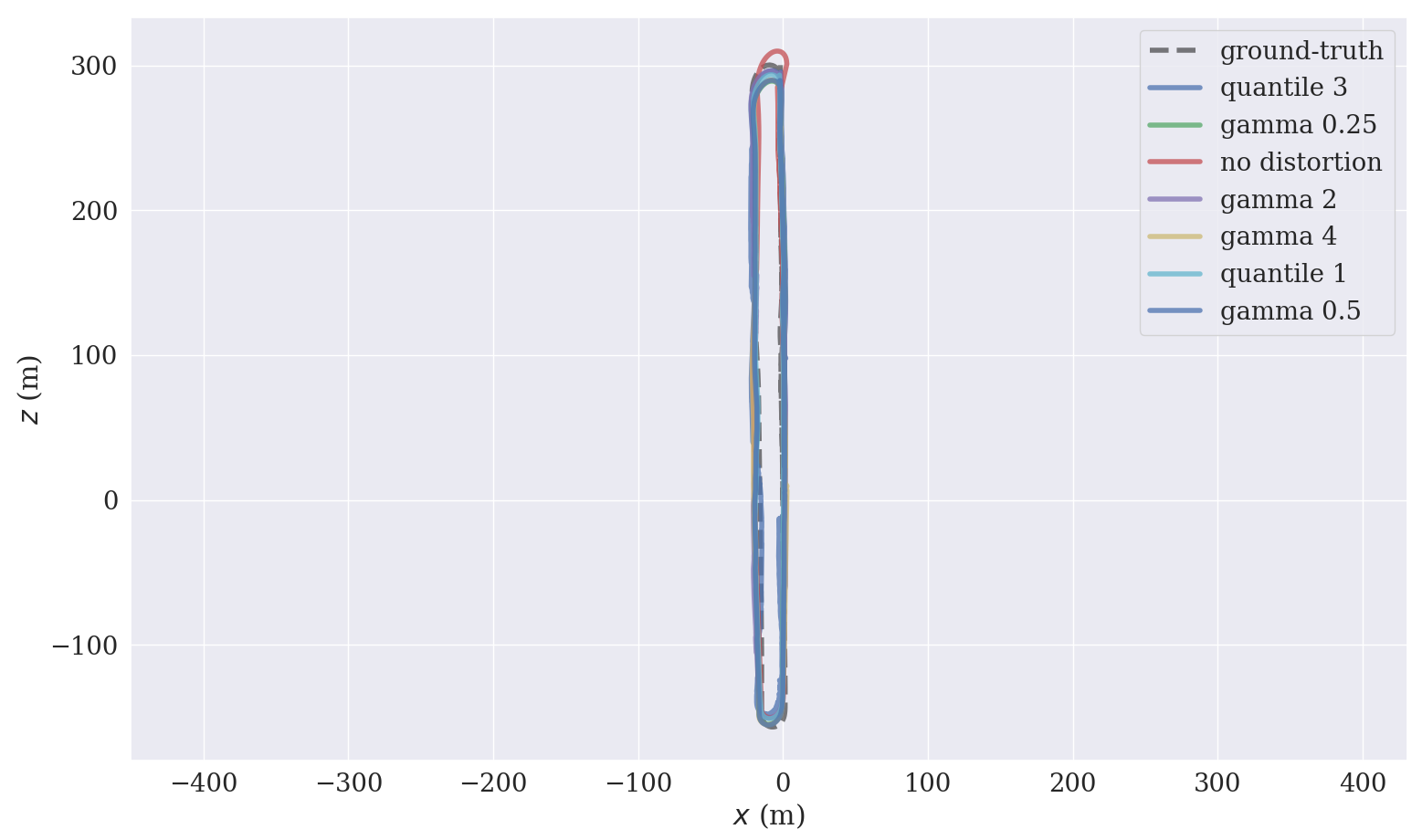}
\label{fig:liftslam-noise-06}}

\subfloat[ORB-SLAM in KITTI 07]{
\includegraphics[width=0.4\textwidth, height=3cm]{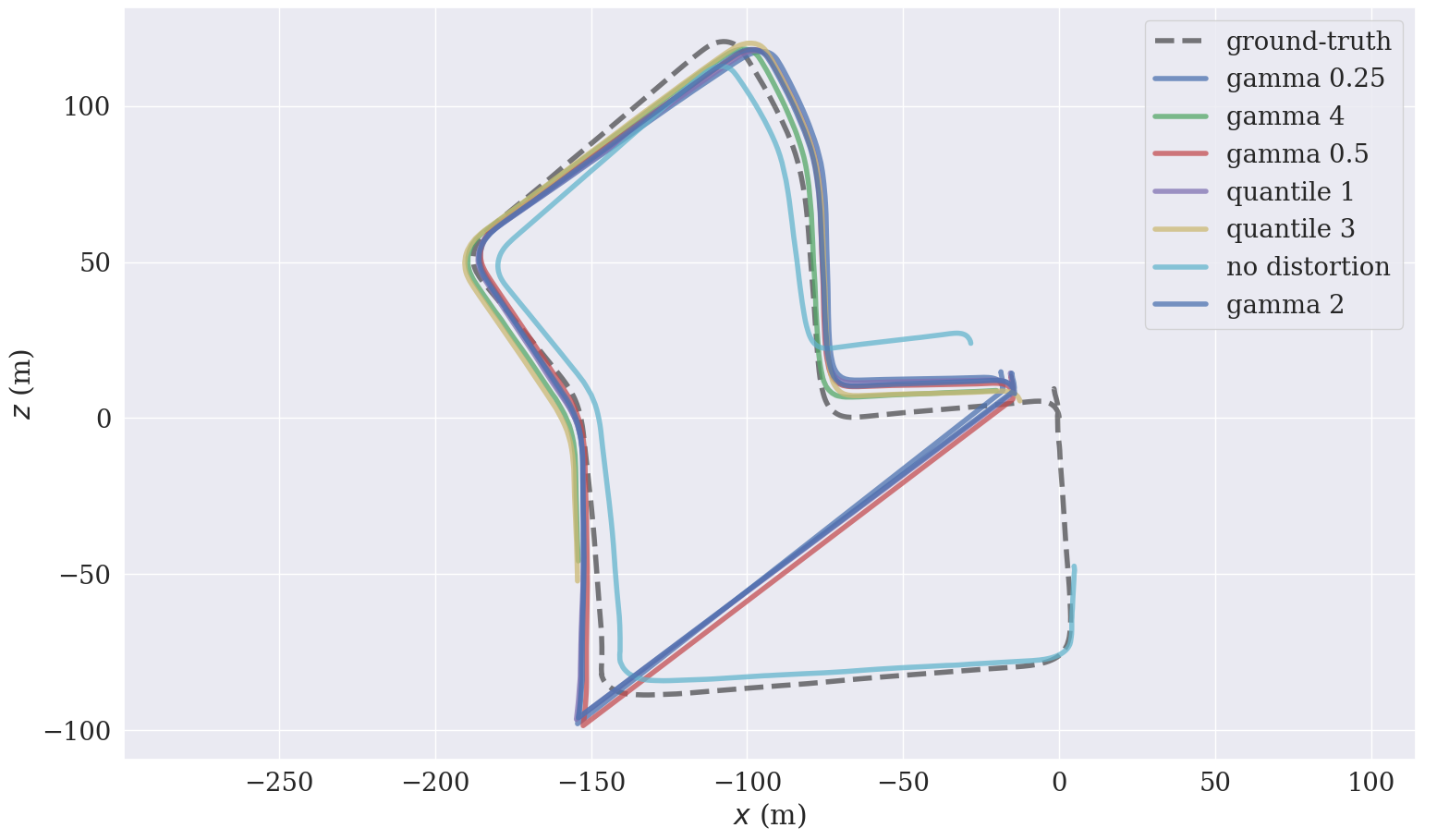}
\label{fig:orbslam-noise-07}}
\qquad
\subfloat[LIFT-SLAM in KITTI 07]{
\includegraphics[width=0.4\textwidth, height=3cm]{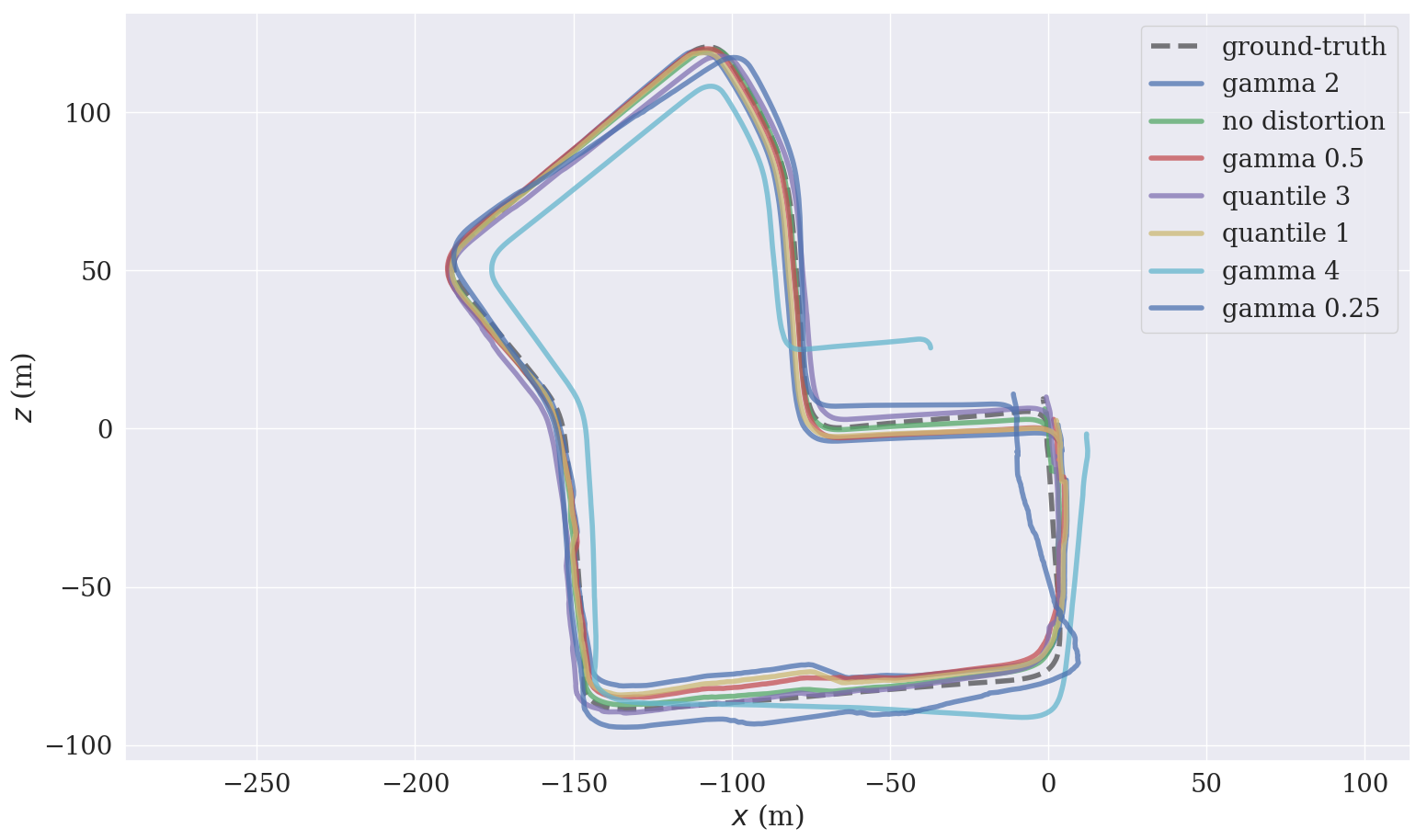}
\label{fig:liftslam-noise-07}}

\subfloat[ORB-SLAM in KITTI 10]{
\includegraphics[width=0.4\textwidth, height=3cm]{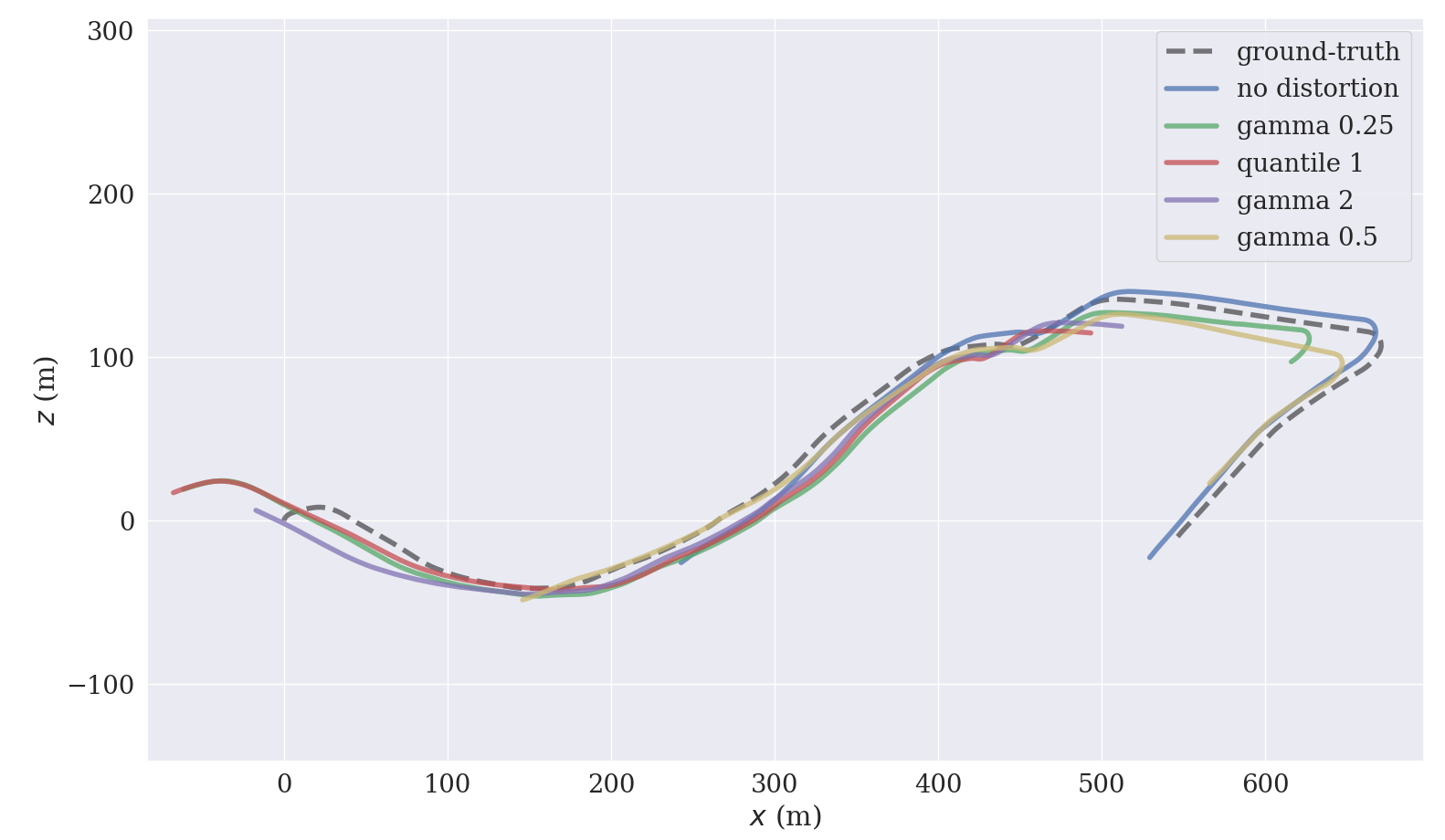}
\label{fig:orbslam-noise-10}}
\qquad
\subfloat[LIFT-SLAM in KITTI 10]{
\includegraphics[width=0.4\textwidth, height=3cm]{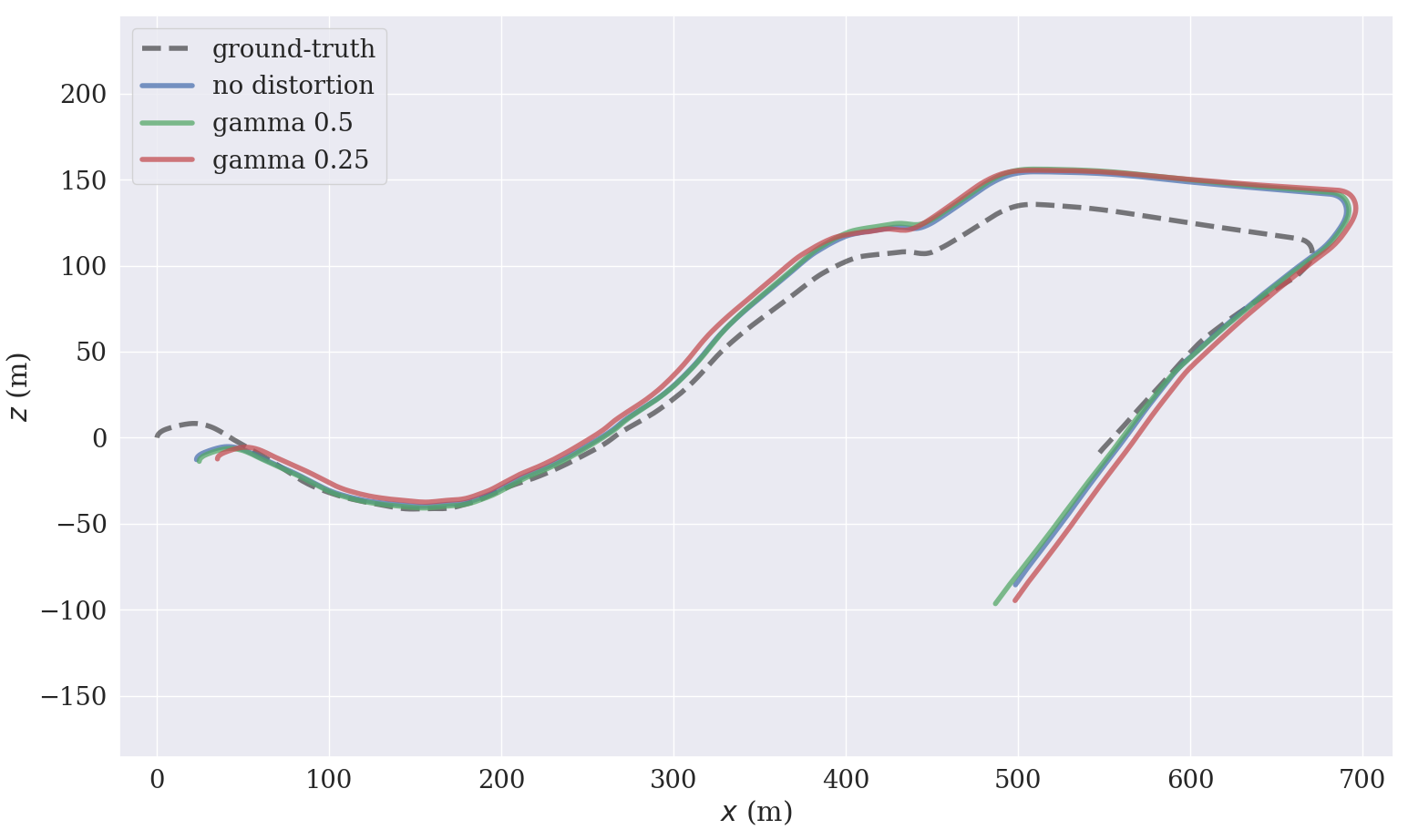}
\label{fig:liftslam-noise-10}}

\subfloat[ORB-SLAM in Euroc MH\_02]{
\includegraphics[width=0.4\textwidth, height=3cm]{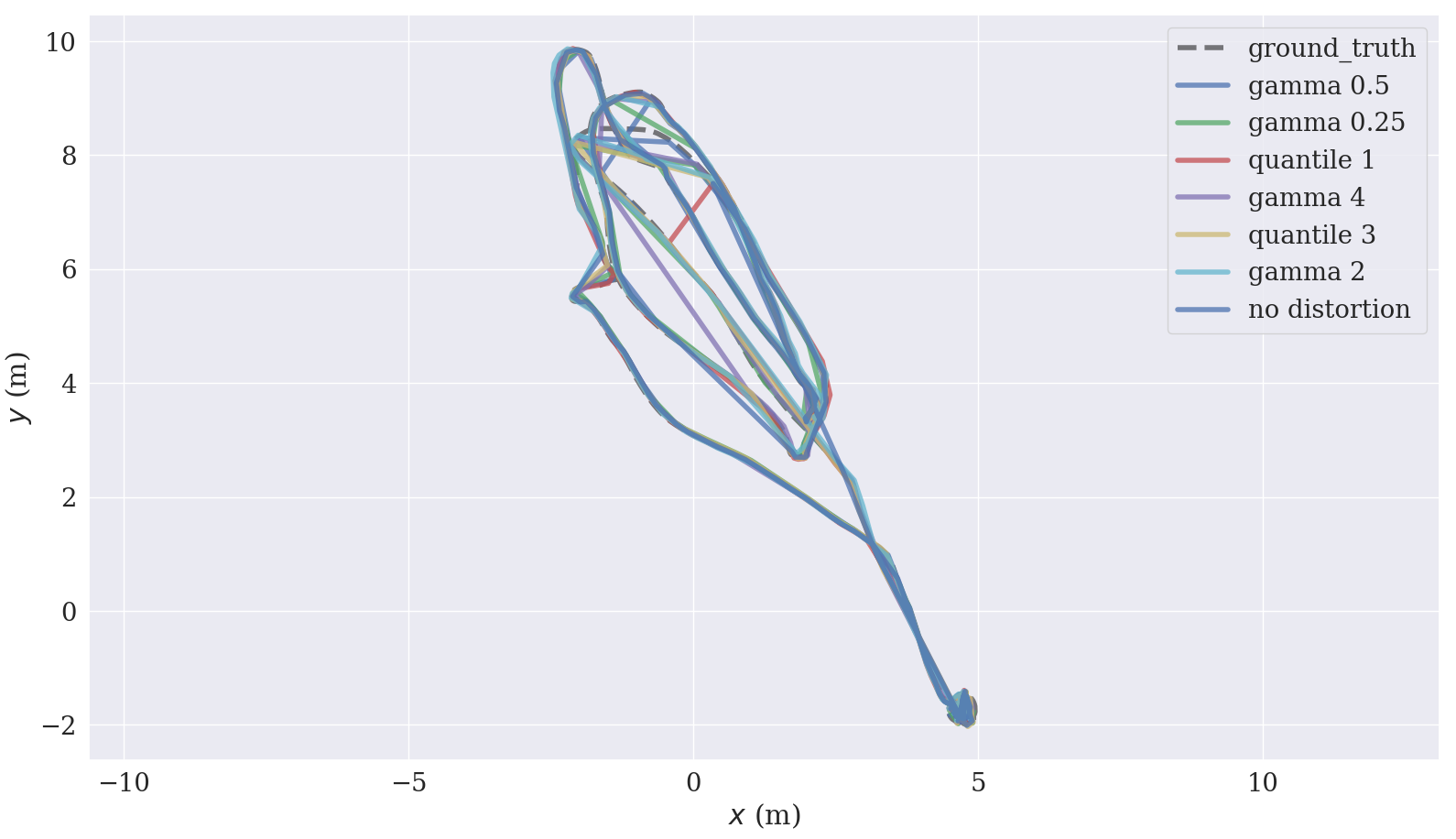}
\label{fig:orbslam-noise-mh02}}
\qquad
\subfloat[LIFT-SLAM in Euroc MH\_02]{
\includegraphics[width=0.4\textwidth, height=3cm]{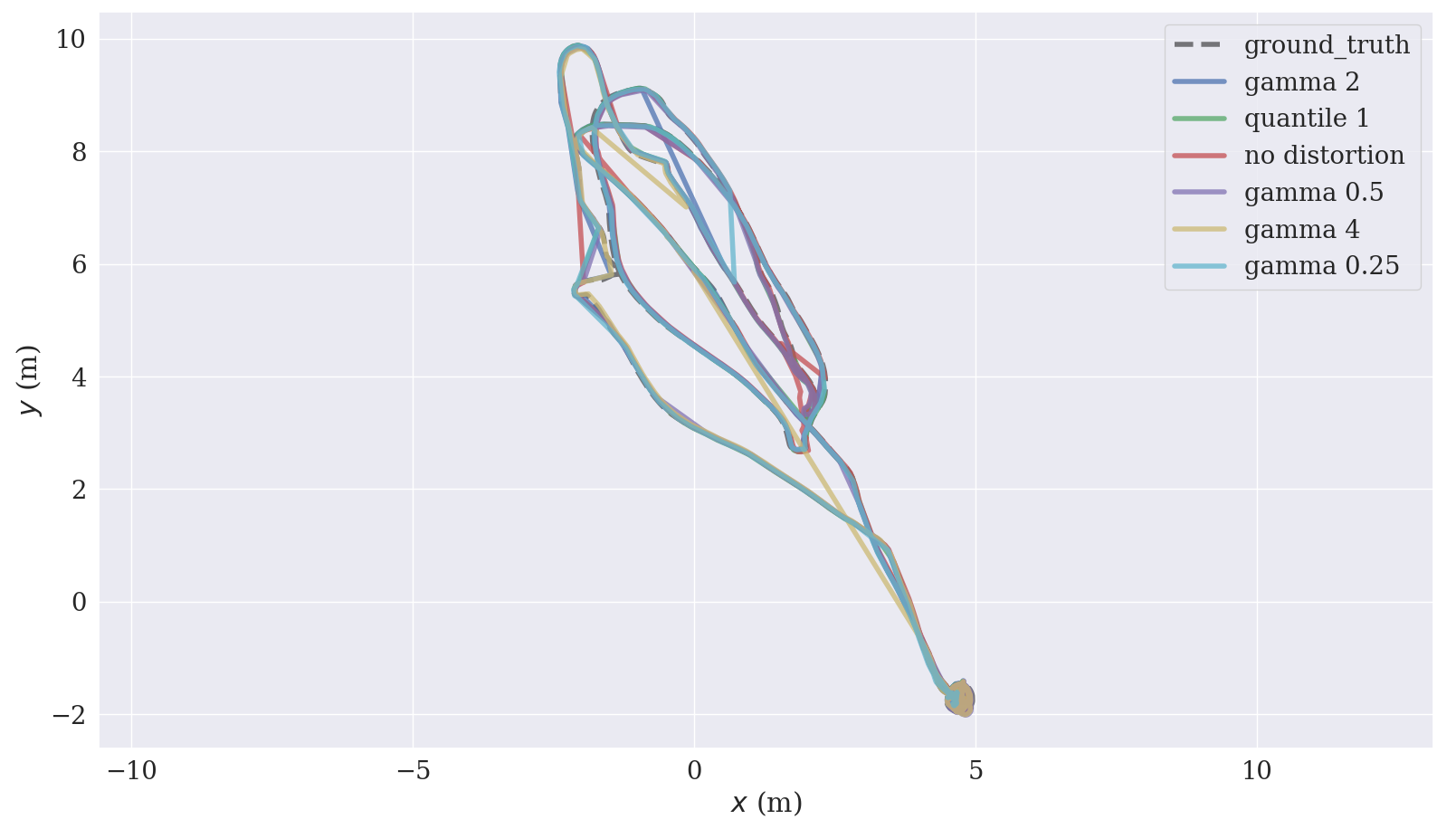}
\label{fig:liftslam-noise-mh02}}

\caption{Qualitative results of the robustness tests.}
\label{fig:all-traj-robustness}
\end{figure}

Figure \ref{fig:all-traj-robustness} shows the comparison of the algorithm's trajectories with each distortion in KITTI and Euroc sequences. In sequences KITTI 03 and KITTI 06, LIFT-SLAM's trajectories were not much affected by distortions (Figures \ref{fig:liftslam-noise-03} and \ref{fig:liftslam-noise-06}). On the other side, the trajectories of ORB-SLAM are worse, especially in sequence KITTI 06 (Figure \ref{fig:orbslam-noise-06}). Furthermore, the trajectories of both algorithms were more affected in KITTI 07 (Figures \ref{fig:orbslam-noise-07} and \ref{fig:liftslam-noise-07}), but ORB-SLAM could not track a considerable part of the trajectory in most scenarios, while LIFT-SLAM could. In MH\_02, both algorithms' trajectories were less affected, but they lost track of the pose and relocalized in some parts of the sequence. Therefore, we can conclude by quantitative and qualitative results that LIFT-SLAM is more robust to the distortions we applied in the sequences. The main reason for this is because the learned features can handle better camera ill exposure, as the datasets used to train and fine-tune the network naturally contain varying illumination.

\subsection{Comparison with literature}

The results obtained have shown that LIFT is capable of improving a traditional VSLAM algorithm. Moreover, transfer learning proved to be a crucial process in our system since it improved our algorithms in different VSLAM problems. The main drawback in our system is that the good performance in large environments depends on loop closure detection, if a loop is not recognized the error increases indefinitely since the drift accumulation is not corrected with pose graph optimization, as in sequences $08$ and $09$.

Furthermore, currently, LIFT-SLAM is slow compared to other state-of-the-art algorithms because we did not optimize the LIFT code for real-time execution. The network does not have many parameters (approximately 290K) when compared to other deep networks. However, the code has some time-consuming operations that are unnecessary for our pipeline. Therefore, the time spent to generate the LIFT descriptors in a GTX 1050 Ti is 36 seconds for images from the KITTI dataset and 35 seconds for images from the Euroc dataset. This code can be improved to perform better and work in real-time (30Hz). For the sake of comparison, in the same machine, the Yolo V3 network \cite{yolov3}, with 61.9M parameters, takes approximately 22 seconds to detect objects in a KITTI image and 21 seconds in a Euroc image. 

We chose Adaptive LIFT-SLAM fine-tuned with Euroc sequences to compare with some results available in the literature. We provide in table \ref{tab:comparison-sota} compared to other algorithms in the KITTI dataset. We selected different monocular VO and VSLAM algorithms to compare with LIFT-SLAM: traditional methods, hybrid methods, and end-to-end methods. In this way, we could compare our results with algorithms that present different characteristics and are trained directly from KITTI images. Unfortunately, there are not many monocular algorithms that evaluate Euroc available in the literature. 

\begin{table}[h]
\begin{threeparttable}

\centering
\resizebox{\textwidth}{!}{\begin{tabular}{|c|c|c|ccccccccccc|}
\hline \textbf{Algorithm} & \textbf{Type} & \textbf{Metric} & \textbf{00} & \textbf{01} & \textbf{02} & \textbf{03} & \textbf{04} & \textbf{05} & \textbf{06} & \textbf{07} & \textbf{08} & \textbf{09} & \textbf{10}\\ \hline

          & &ATE (m) & 8.06 & X & 40.04 & \textbf{2.23} & \textbf{0.51} & 13.55 & 30.38 & \textbf{3.63} & 184.43 & 59.62 & 29.87 \\
LIFT-SLAM & Hybrid & $RPE_{trans}$ (\%) & \textbf{3.18} & X & 8.73 & \textbf{1.46} & \textbf{2.22} & 6.09 & 12.24 & \textbf{2.42} & 47.10 & 19.91 & 9.72 \\
          & &$RPE_{rot}$ (deg/m) & 2.99 & X & 2.49 & \textbf{0.34} & \textbf{0.48} & 3.11 & 2.91 & 4.02 & \textbf{2.02} & 2.14 & \textbf{2.24}  \\  \hline

           && ATE (m) & 11.54 & X & X & 15.13& 4.29 & \textbf{7.74} & 20.26 & 13.47 & \textbf{39.51} & 49.67 & \textbf{19.94}\\
ORB-SLAM\tnote{*}   & Traditional  & $RPE_{trans}$ (\%) & 4.46 & X & X & 9.75 & 3.71 & 3.35 & 8.11 & 7.43 & \textbf{12.16} & 26.51 & 8.65 \\
            && $RPE_{rot}$ (deg/m) & 3.28 & X & X & 2.78 & 2.15 & 3.57 & 2.88 & 3.58 &3.05 & 11.13 & 3.62 \\ \hline

          & &ATE (m) & \textbf{5.33} & X & \textbf{21.28} & 1.51 & 1.62 & 4.85 & \textbf{12.34} & 2.26 & 46.68 & \textbf{6.62} & 8.80 \\
ORB-SLAM \cite{orb-slam} & Traditional & $RPE_{trans}$ (\%) &-&-&-&-&-&-&-&-&-&-&- \\
                        & &$RPE_{rot}$ (deg/m) &-&-&-&-&-&-&-&-&-&-&- \\ \hline

                     & &ATE (m)&-&-&-&-&-&-&-&-&-&-&- \\ 
DeepVO\cite{deep-vo}\tnote{**}& End-to-end & $RPE_{trans}$ (\%)&-&-&-& 8.49 & 7.19 & 2.62 & 5.42 & 3.91 &-&-& \textbf{8.11} \\
                    & &$RPE_{rot}$ (deg/m)&-&-&-& 6.89 & 6.97 & 3.61 & 5.82 & 4.60 &-&-& 8.83 \\\hline


                  &  & ATE (m) &-&-&-&-&-&-&-&-&-&-&- \\ 
NeuralBundler \cite{pose-graph-optimization}& Hybrid & $RPE_{trans}$ (\%) & 3.24 & - & \textbf{4.85} & - & - & \textbf{1.83} & \textbf{2.74} & 3.53 &-& \textbf{6.23} &- \\
                    && $RPE_{rot}$ (deg/m) & \textbf{1.35} & - & \textbf{1.60} & - & - &\textbf{0.7} & \textbf{2.6} & \textbf{2.02} & - &\textbf{2.11} & - \\\hline
                    
\end{tabular}}
        \begin{tablenotes}
            \item[*] {\footnotesize Our executions.}
            \item[**]{\footnotesize  Only VO.}
        \end{tablenotes}
\end{threeparttable}

\caption{Comparison of LIFT-SLAM with results from monocular VO/VSLAM algorithms available in the literature. We fill with "X" results that are unavailable due to tracking failure and with "-" results that were not given by the authors.}
\label{tab:comparison-sota}
\end{table}

Table \ref{tab:comparison-sota} shows that LIFT-SLAM obtained the smallest error in sequences 00, 03, 04, 07, 08, and 10. Additionally, we can verify that the end-to-end approach is not as accurate as traditional and hybrid approaches. It is important to mention that the LIFT-SLAM version used in this evaluation was not finetuned with any KITTI sequence, while DeepVO \cite{deep-vo} and NeuralBundler \cite{pose-graph-optimization} approaches were trained with some KITTI sequences,  typically indicating an overfit over this dataset. Nevertheless, we still present competitive results without overfitting in the dataset.

\begin{figure}
\centering
\subfloat[LIFT-SLAM and DeepVO trajectories in KITTI 03.]{
\includegraphics[width=0.4\textwidth, height=3cm]{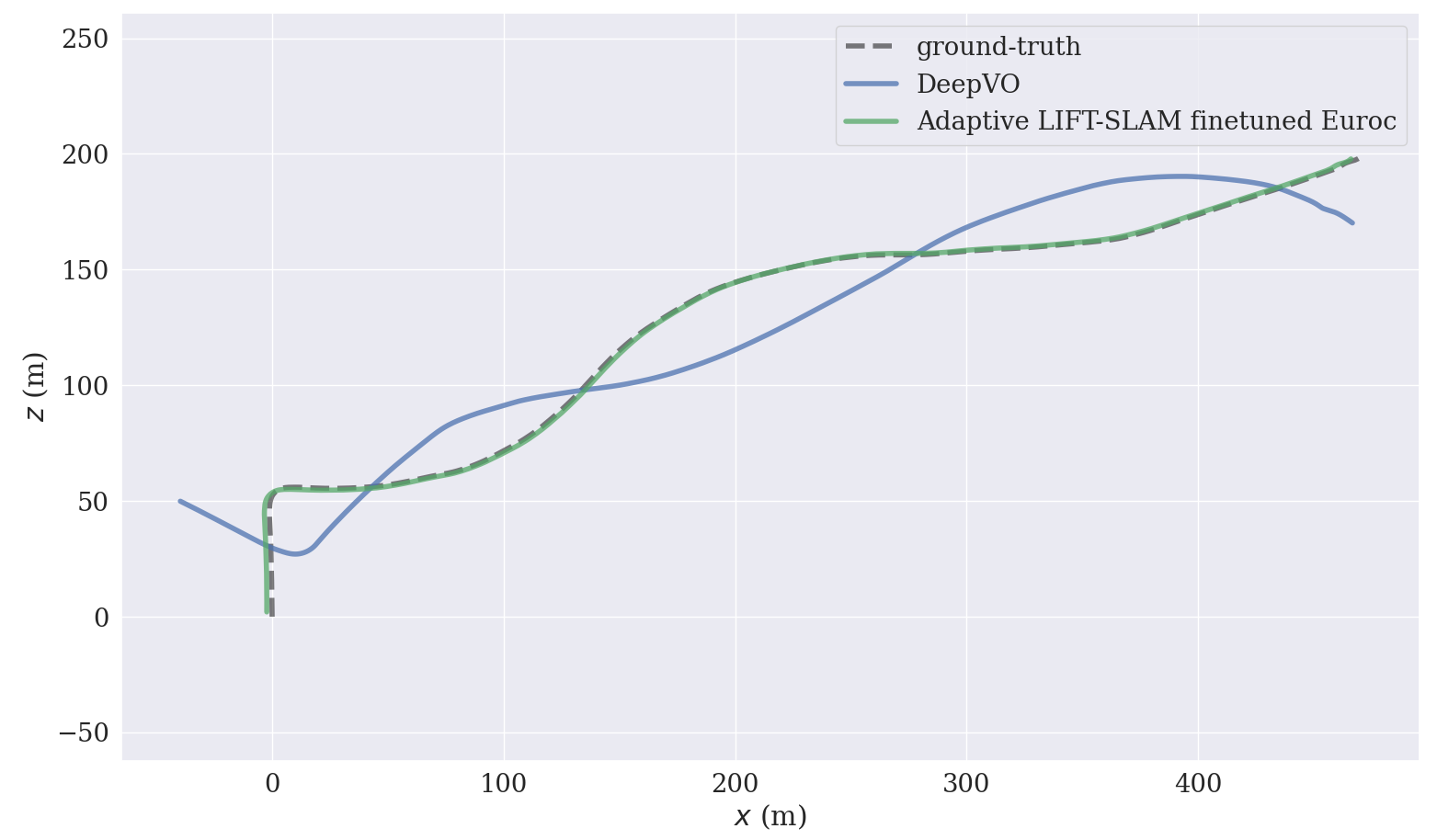}
\label{fig:deepvo-03}}
\qquad
\subfloat[LIFT-SLAM and DeepVO trajectories in KITTI 04.]{
\includegraphics[width=0.4\textwidth, height=3cm]{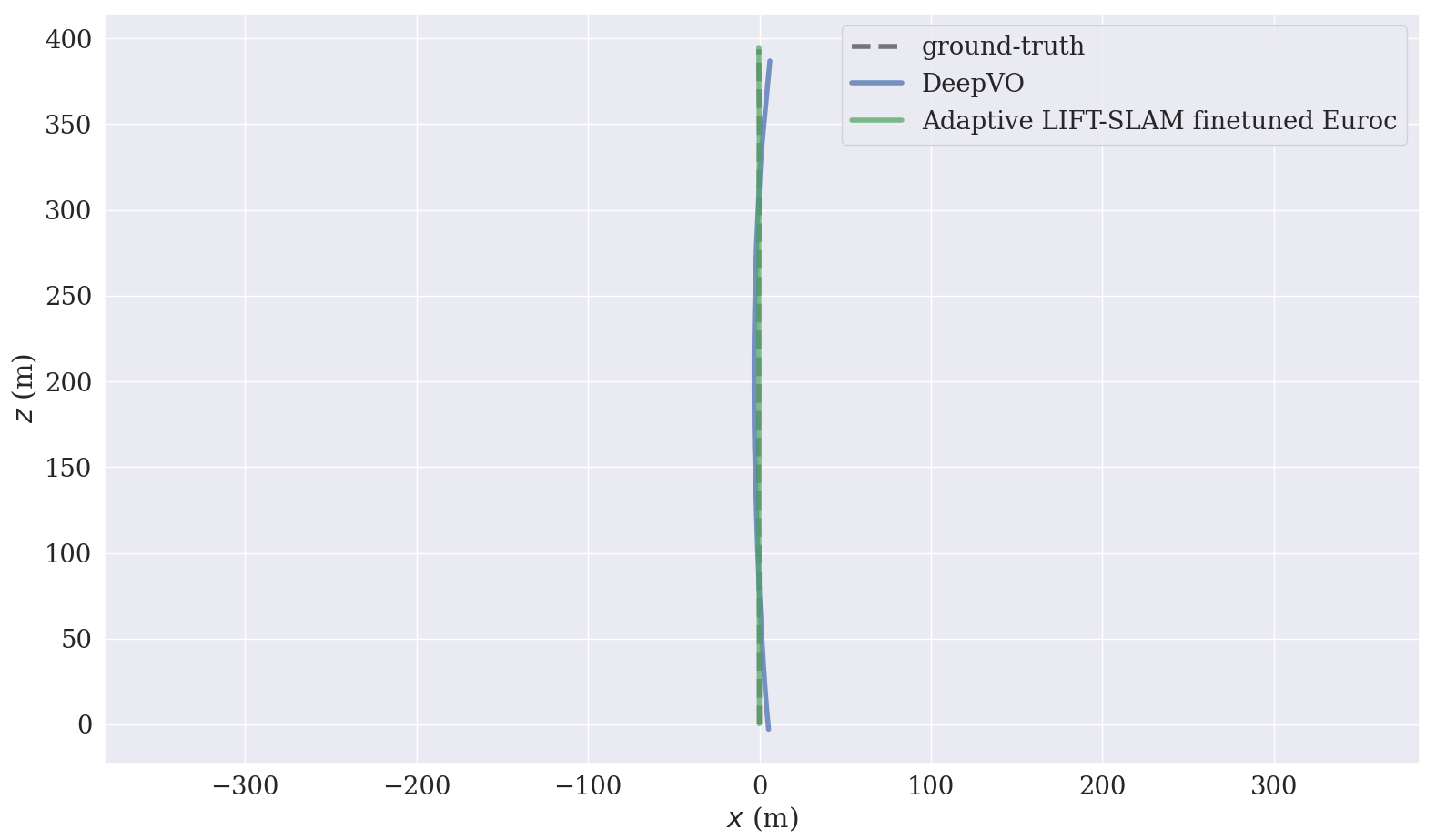}
\label{fig:deepvo-04}}

\subfloat[LIFT-SLAM and DeepVO trajectories in KITTI 05.]{
\includegraphics[width=0.4\textwidth, height=3cm]{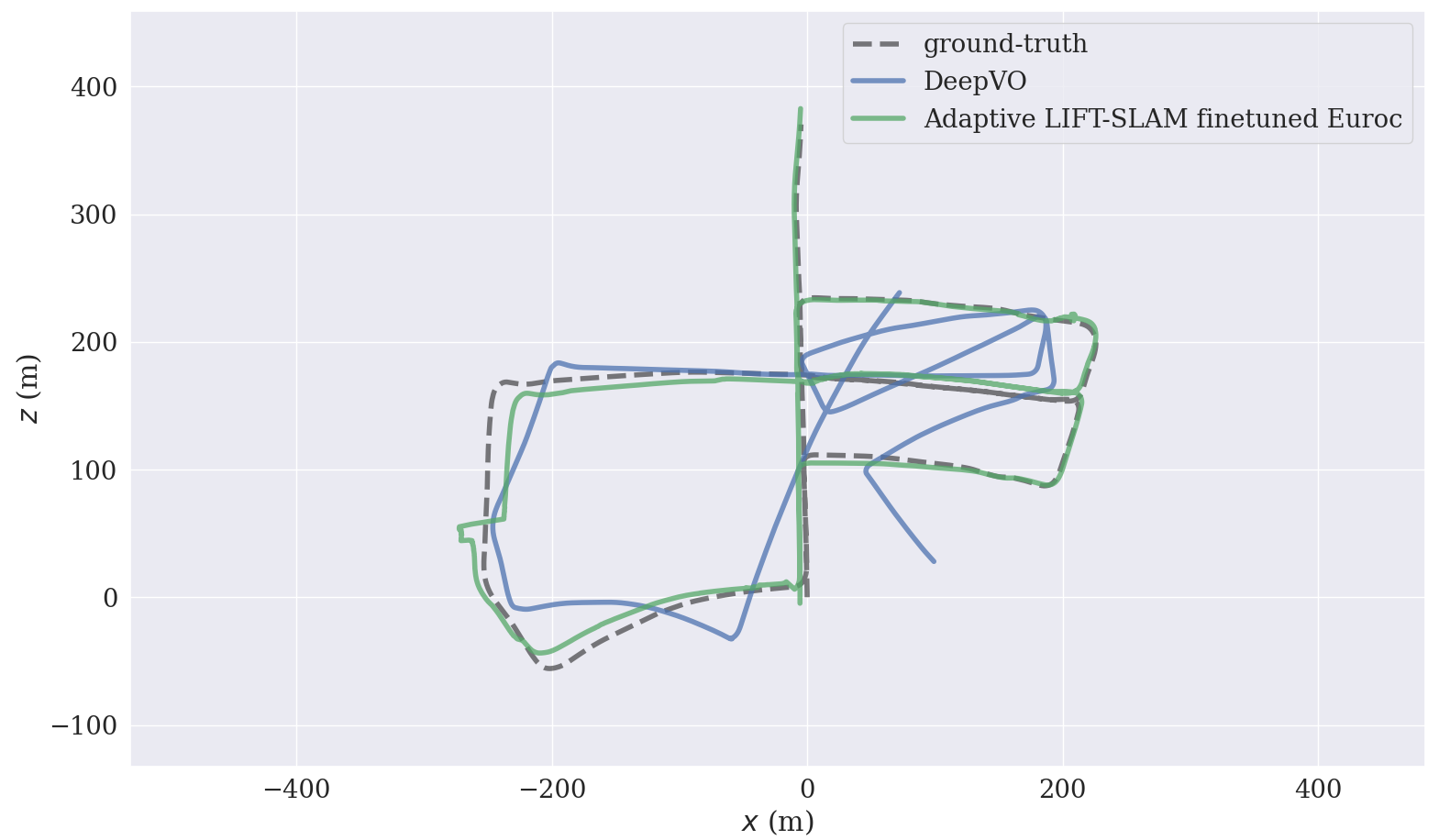}
\label{fig:deepvo-05}}
\qquad
\subfloat[LIFT-SLAM and DeepVO trajectories in KITTI 06.]{
\includegraphics[width=0.4\textwidth, height=3cm]{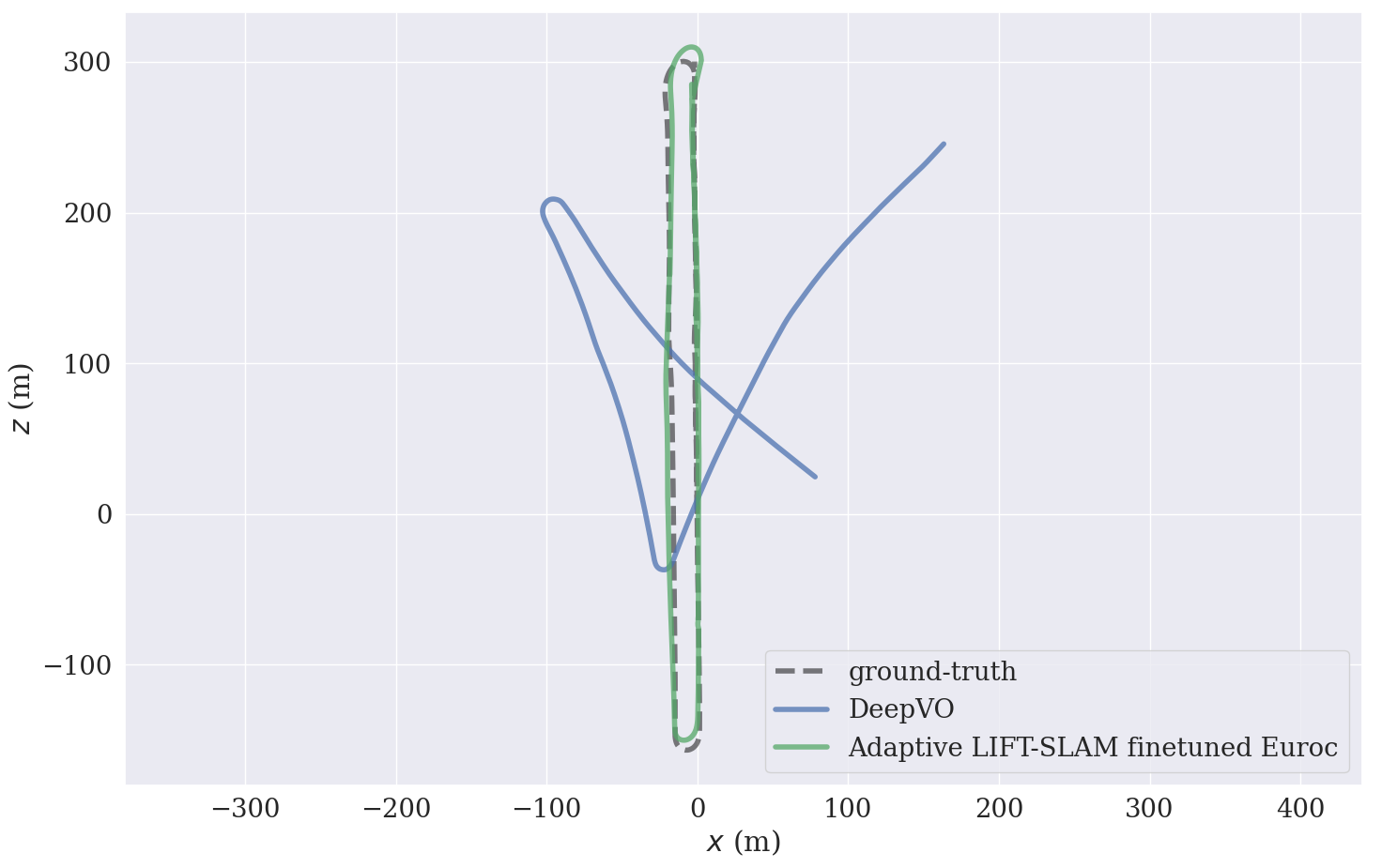}
\label{fig:deepvo-06}}

\subfloat[LIFT-SLAM and DeepVO trajectories in KITTI 07.]{
\includegraphics[width=0.4\textwidth, height=3cm]{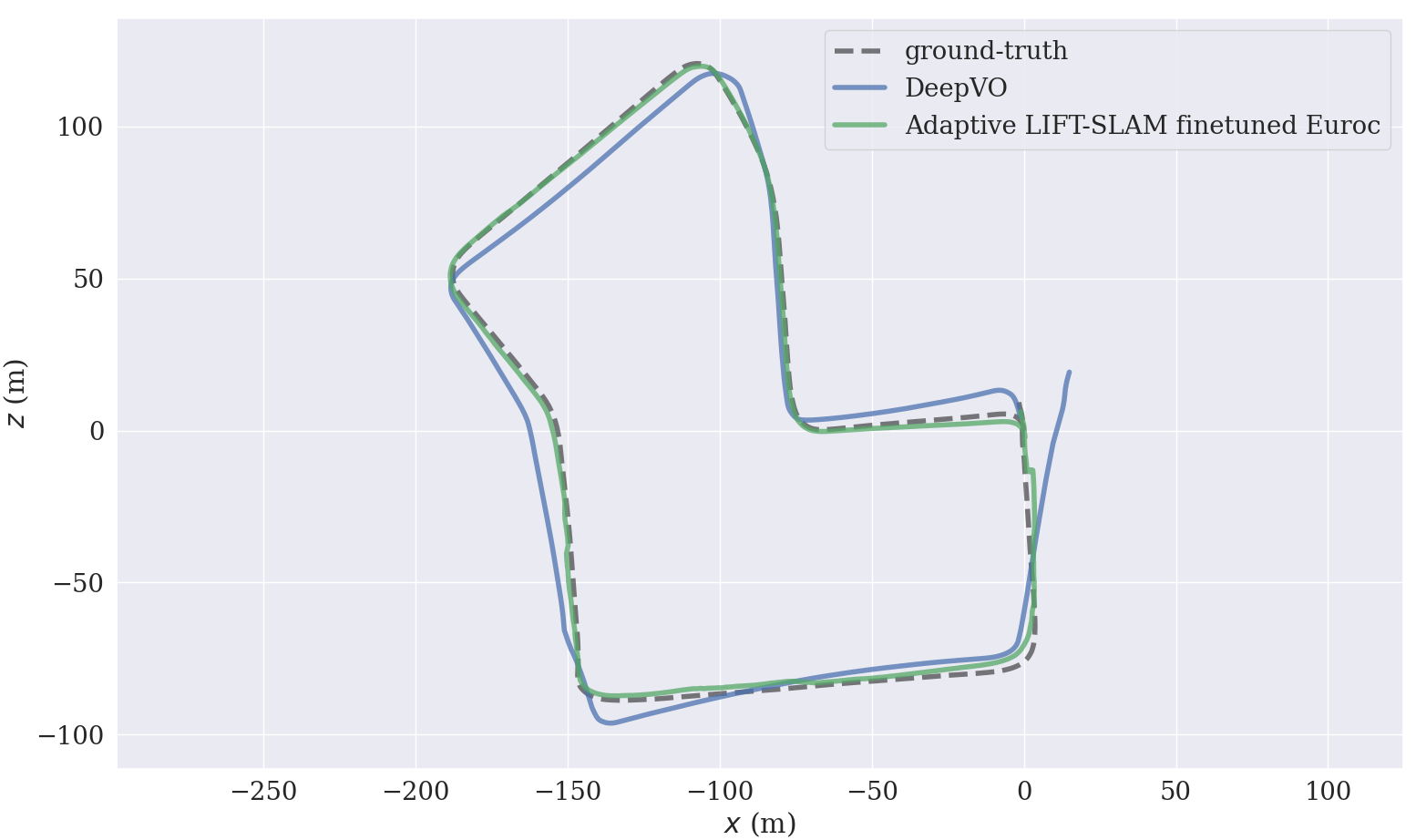}
\label{fig:deepvo-07}}
\qquad
\subfloat[LIFT-SLAM and DeepVO trajectories in KITTI 10.]{
\includegraphics[width=0.4\textwidth, height=3cm]{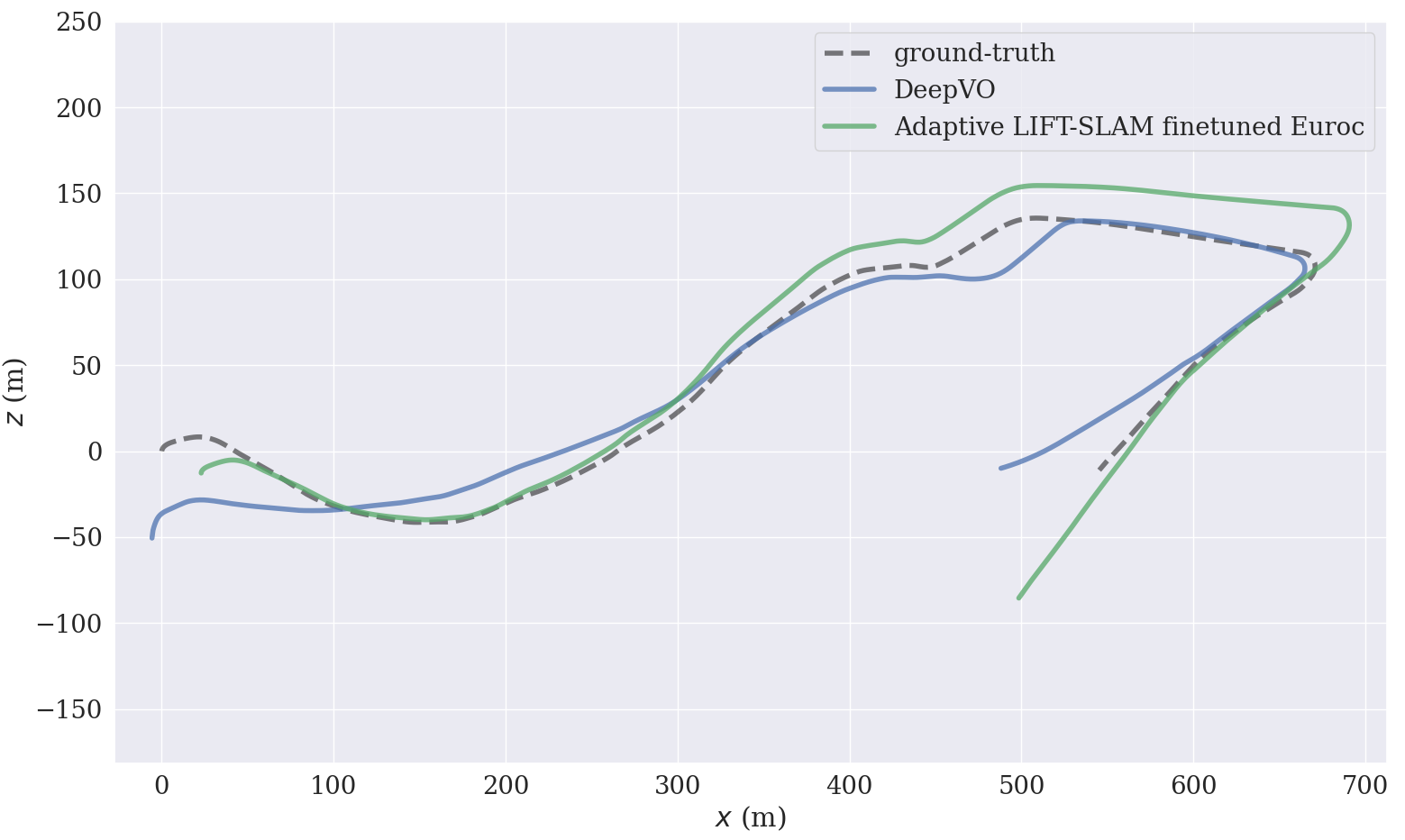}
\label{fig:deepvo-10}

}

\caption{Qualitative comparison with DeepVO trajectories we generated in KITTI dataset. The LIFT-SLAM version used in this comparison is the Adaptive finetuned with Euroc sequences.}
\label{fig:deepvo-vs-ours}
\end{figure}

Figure \ref{fig:deepvo-vs-ours} shows a qualitative comparison between our algorithm and DeepVO \cite{deep-vo}. The DeepVO trajectories were generated by us, based on the model trained by the unofficial PyTorch implementation available at \cite{deepvo-github}. In most of the sequences, LIFT-SLAM trajectories are closer to the ground-truth. It is important to remark that DeepVO has no loop-closure detection. However, LIFT-SLAM performed better even in sequences without a loop, such as 03 (Fig. \ref{fig:deepvo-03}) and 04 (Fig. \ref{fig:deepvo-04}). Unfortunately, there are no hybrid methods for monocular VSLAM with open code available, so we could not evaluate these algorithms' qualitative results.

\section{Conclusion}
\label{sec:conclusion}

In this work, we successfully apply a deep neural network in the front-end of a traditional monocular visual SLAM algorithm. This approach showed that it is possible to improve VSLAM algorithms' performance with learned feature extraction and description. We also showed that transfer learning could be used to fine-tune these networks with VO/VSLAM datasets to improve the entire system's performance on cross-datasets. Moreover, we successfully created a method to adapt the matching thresholds while executing the VO pipeline, depending on the number of outliers. This method allowed us to eliminate the fixed values of the matching thresholds without requiring dataset fine-tuning. All of these methods allowed us to evaluate five variations of LIFT-SLAM: LIFT-SLAM fine-tuned with KITTI sequences, LIFT-SLAM fine-tuned with Euroc sequences, Adaptive LIFT-SLAM, Adaptive LIFT-SLAM fine-tuned with KITTI sequences and Adaptive LIFT-SLAM fine-tuned with Euroc sequences. 

We also proposed a set of experiments to evaluate the robustness of VSLAM algorithms. With these experiments, we showed that our hybrid VSLAM algorithm is more robust than a traditional VSLAM algorithm without losing accuracy, such as end-to-end deep learning-based algorithms. Results demonstrate that the proposed system can operate in different environments (indoors and outdoors) while improving its results with an artificial distortion applied to the images (gamma power transformation and quantile-based truncation). This fact indicates that a selection of the learned features could improve the algorithm' performance. Therefore, in future work, we plan to add an attention-based mechanism to select the best features for VSLAM.
\section{Acknowledgments}
This work was supported by the Brazilian National Council for Scientific and Technological Development (CNPq) and by the company Quinto Andar.




\bibliographystyle{elsarticle-num} 
\bibliography{references}



\end{document}